\documentclass{article} % For LaTeX2e
\usepackage{iclr2023_conference,times}

\usepackage{amsmath,amsfonts,bm}

\def\eqref#1{equation~\ref{#1}}
\def\1{\bm{1}}

\DeclareMathAlphabet{\mathsfit}{\encodingdefault}{\sfdefault}{m}{sl}
\SetMathAlphabet{\mathsfit}{bold}{\encodingdefault}{\sfdefault}{bx}{n}

\usepackage{times}
\usepackage{epsfig}
\usepackage{graphicx} 
\usepackage[utf8]{inputenc} % allow utf-8 input
\usepackage[T1]{fontenc}    % use 8-bit T1 fonts
\usepackage{xcolor}         % colors
\usepackage{amsmath}
\usepackage{amsfonts}
\usepackage{array}
\usepackage{booktabs}
\usepackage{multirow}
\usepackage{makecell}
\usepackage{float} 
\usepackage{xspace}
\usepackage{caption}

\usepackage{algorithm}
\usepackage{algpseudocode}

\usepackage[pagebackref=true,breaklinks=true,colorlinks,bookmarks=false,hypertexnames=false]{hyperref}
\usepackage[capitalize]{cleveref}
\crefname{section}{Sec.}{Secs.}
\Crefname{section}{Section}{Sections}
\Crefname{table}{Table}{Tabs.}
\crefname{table}{Table}{Tabs.}
\usepackage{animate}

\renewcommand{\paragraph}[2][\ ]{\vspace{4pt}\noindent{\bf #2#1}}

\newcommand*{\newcite}[1]{~\cite{#1}}
\newcommand{\norm}[1]{\left\lVert#1\right\rVert}
\newcommand{\myquote}[1]{``#1''}
\definecolor{RWTHblue}{RGB}{0,85,169}%{0,84,159}
\definecolor{BMVCblue}{RGB}{0, 0, 102}
\newcommand{\textblue}[1]{\textcolor{RWTHblue}{#1}}

\newcommand{\jsd}{binding}
\newcommand{\animalScene}{Animal-Scene}
\newcommand{\colorScene}{Color-Obj-Scene}
\newcommand{\tifaCOCO}{COCO-Subject}
\newcommand{\tifaAttr}{COCO-Attribute}
\newcommand{\multiobject}{Multi-Object}
\newcommand{\ours}{Divide \& Bind}
\newcommand{\AandE}{A\&E}

\title{Divide \& Bind Your Attention for Improved Generative Semantic Nursing}

\author{Yumeng Li$^{1,2}$ \  \  Margret Keuper$^{2,3}$\ \ Dan Zhang$^{1,4}$ \ \ Anna Khoreva$^{1,4}$ \\
$^1$ Bosch Center for Artificial Intelligence  \\ $^2$ University of Mannheim \\
$^3$ Max Planck Institute for Informatics \\ 
$^4$ University of T\"ubingen \\
\texttt{\{yumeng.li, dan.zhang2, anna.khoreva\}@de.bosch.com} \\
\ \ \texttt{keuper@uni-mannheim.de}
}

\iclrfinalcopy % Uncomment for camera-ready version, but NOT for submission.
\begin{document}
\maketitle

\begin{abstract}
Emerging large-scale text-to-image generative models, e.g., Stable Diffusion (SD), have exhibited overwhelming results with high  fidelity.  Despite the magnificent progress, current state-of-the-art models still struggle to generate images fully adhering to the input prompt. 
Prior work,  Attend \& Excite, has introduced the concept of Generative Semantic Nursing (GSN), aiming to optimize cross-attention during inference time to better incorporate the semantics. 
It demonstrates promising results in generating simple prompts, e.g., \myquote{a cat and a dog}.
However, its efficacy declines when dealing with more complex prompts, and it does not explicitly address the problem of improper attribute binding.
To address the challenges posed by complex prompts or scenarios involving multiple entities and to achieve improved attribute binding, we propose {\ours}. We introduce two novel loss objectives for GSN: a novel attendance loss and a binding loss.
Our approach stands out in its ability to faithfully synthesize desired objects with improved attribute alignment from complex prompts and exhibits superior performance across multiple evaluation benchmarks. 
More videos and updates can be found on the \href{https://sites.google.com/view/divide-and-bind}{project page}, and \href{https://github.com/boschresearch/Divide-and-Bind}{source code} is available. 
\end{abstract}

\begin{figure*}[h]
\begin{centering}
\setlength{\tabcolsep}{0.0em}
\renewcommand{\arraystretch}{0}
\par\end{centering}
\begin{centering}
%\vspace{-1.8em}
\hfill{}%
	\begin{tabular}{
 @{\hspace{-0.25em}}c
 @{\hspace{0.1em}}c@{\hspace{0.4em}}c
 @{\hspace{0.6em}}c@{\hspace{0.4em}}c
 @{\hspace{0.6em}}c@{\hspace{0.4em}}c
 }
	\centering
    &
    \multicolumn{2}{c}{\begin{tabular}{c}
    \small 
    \myquote{A \textblue{train} driving down the\\ 
    \small  \textblue{tracks}  under a \textblue{bridge} }\end{tabular}} 
    & 
    \multicolumn{2}{c}{\begin{tabular}{c} \small\myquote{\textblue{Ironman} cooking in\\ \small the  kitchen  with a \textblue{dog} }\end{tabular}} 
    &
    \multicolumn{2}{c}{\begin{tabular}{c} 
    \small  \myquote{Three \textblue{geese} floating \\
    \small in the  middle of a river 
    }
    \end{tabular}} 
	%\vspace{0.02cm} 
    %% Stable Diffsuion
    \tabularnewline
    \multirow{1}{*}{\rotatebox{90}{
        \hspace{4.5em}
        \begin{tabular}{c}
        \small Stable\\ \small Diffusion 
        \end{tabular}
        \hspace{-4.5em}
    }} 
    &
	\includegraphics[width=0.14\linewidth]{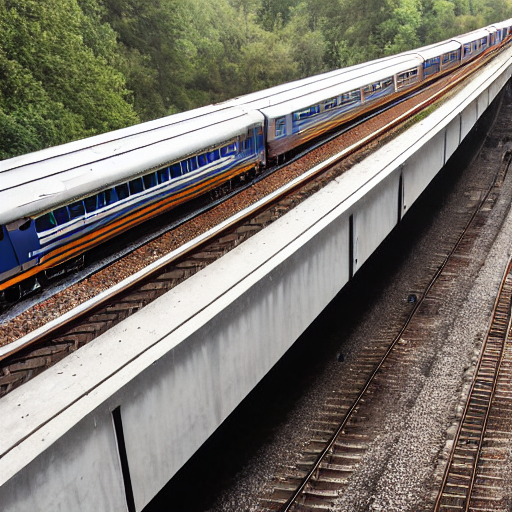}
	      &
	\includegraphics[width=0.14\linewidth]{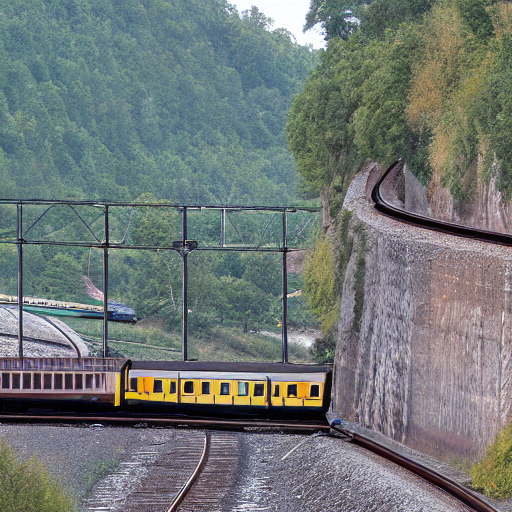}
	    & 
	\includegraphics[width=0.14\linewidth]{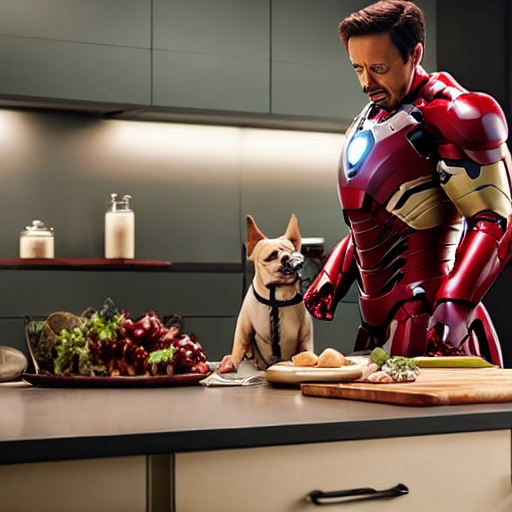}
        & 
    \includegraphics[width=0.14\linewidth]{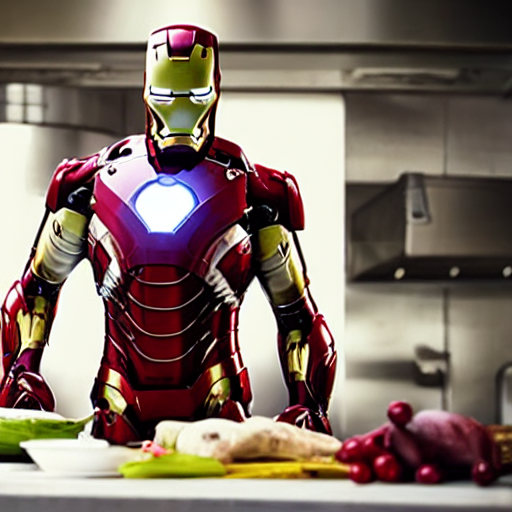}
	      & 
	\includegraphics[width=0.14\linewidth]{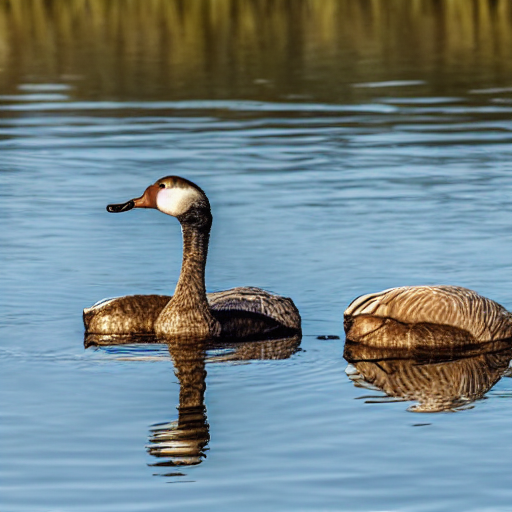}
	    & 
	\includegraphics[width=0.14\linewidth]{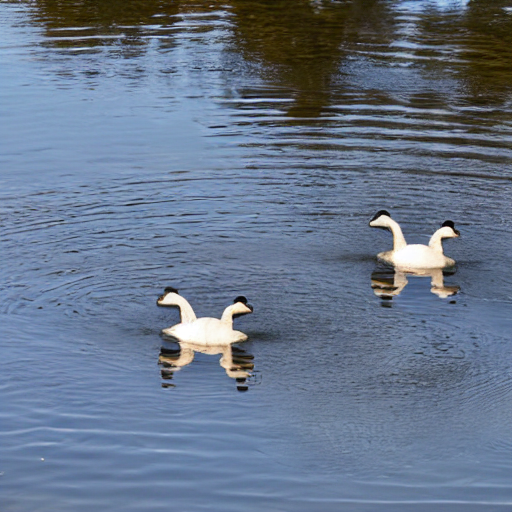}
	 %% Attend and Excite
  \tabularnewline
  \multirow{1}{*}{\rotatebox{90}{
        \hspace{4.5em}
        \begin{tabular}{c}
        \small Attend \& \\ \small Excite
        \end{tabular}
        \hspace{-4.5em}
    }} 
    &
    \includegraphics[width=0.14\linewidth]{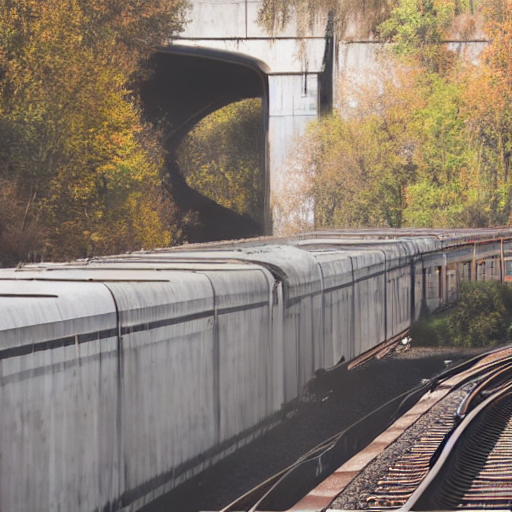}
	      &
	\includegraphics[width=0.14\linewidth]{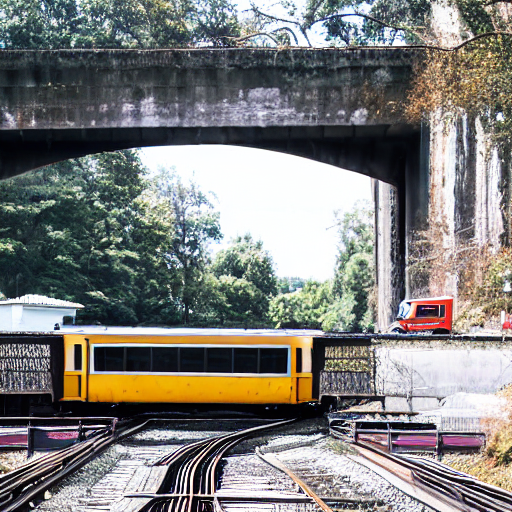}
	    & 
	\includegraphics[width=0.14\linewidth]{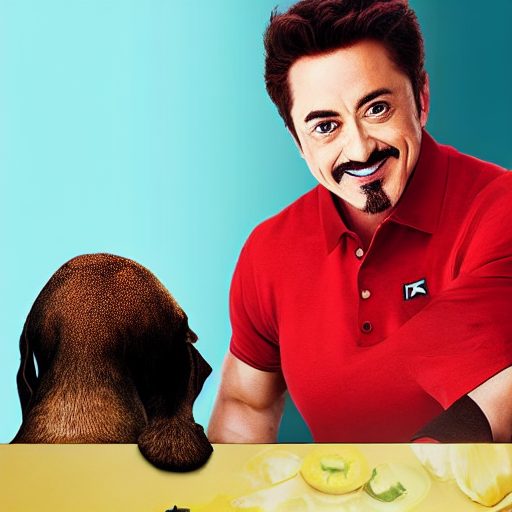}
        & 
    \includegraphics[width=0.14\linewidth]{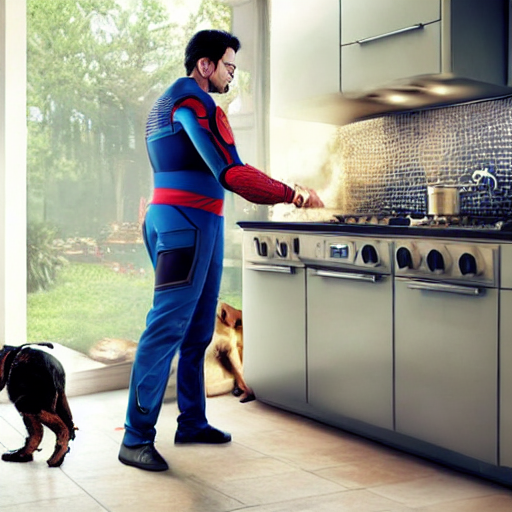}
	      & 
	\includegraphics[width=0.14\linewidth]{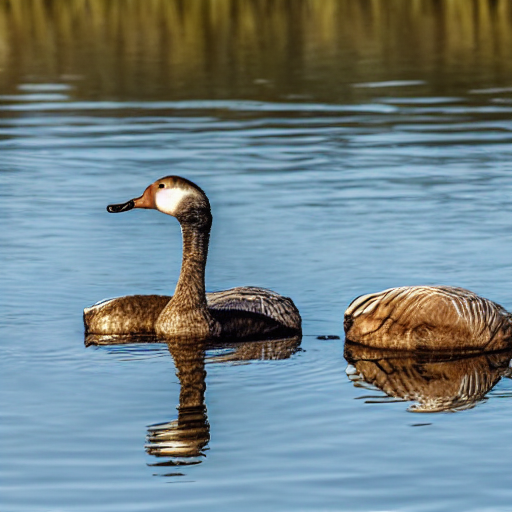}
	    & 
	\includegraphics[width=0.14\linewidth]{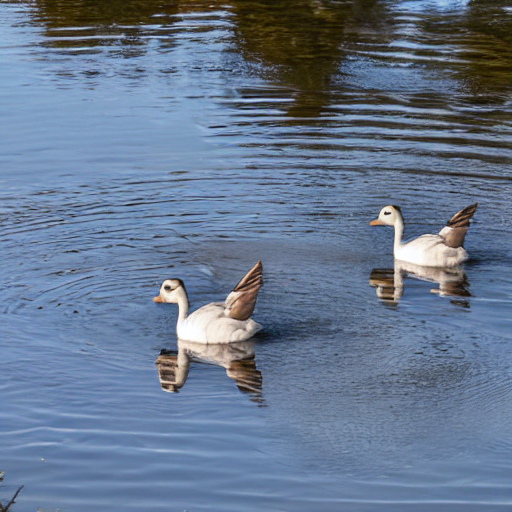}
	 \tabularnewline
  %% Ours
    \multirow{1}{*}{\rotatebox{90}{
        \hspace{4.4em}
        \begin{tabular}{c}
        \small \textbf{Divide \&} \\ \small \textbf{Bind} 
        \end{tabular}
        \hspace{-4.4em}
    }} 
    &
	\includegraphics[width=0.14\linewidth]{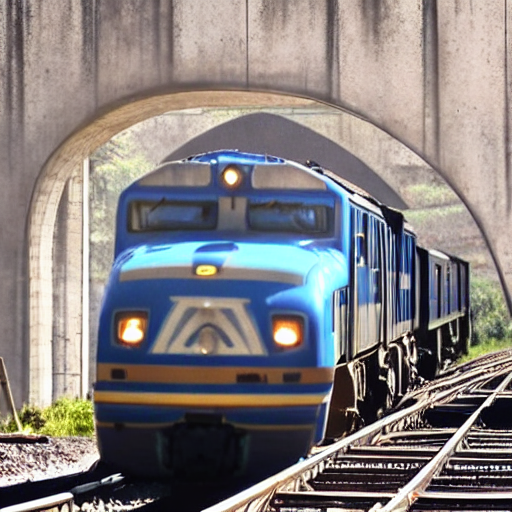}
	      &
	\includegraphics[width=0.14\linewidth]{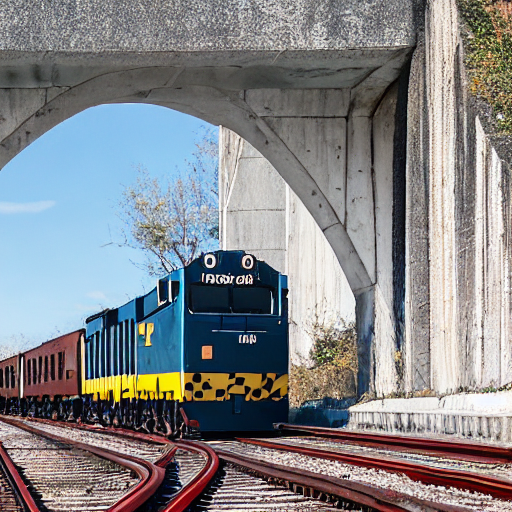}
	    & 
	\includegraphics[width=0.14\linewidth]{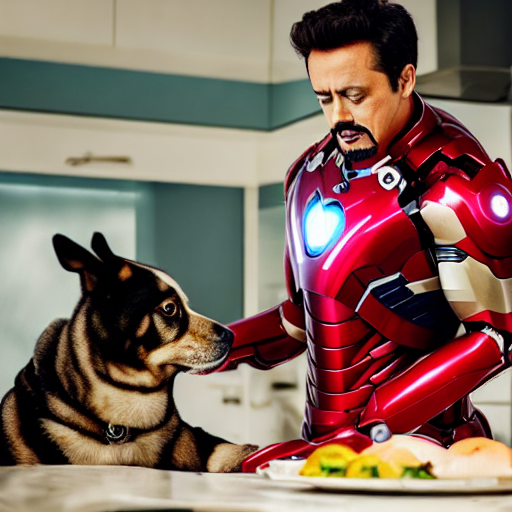}
        & 
    \includegraphics[width=0.14\linewidth]{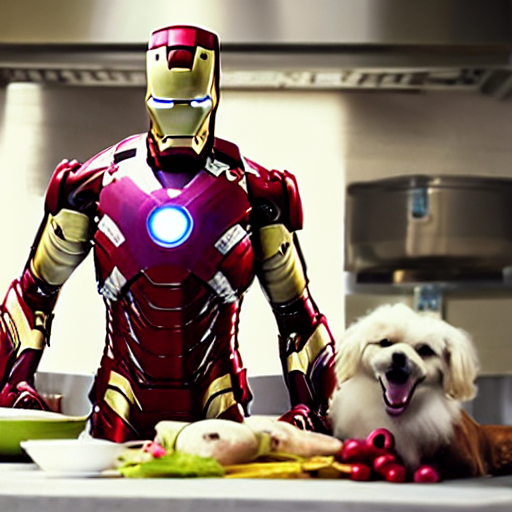}
	      & 
	\includegraphics[width=0.14\linewidth]{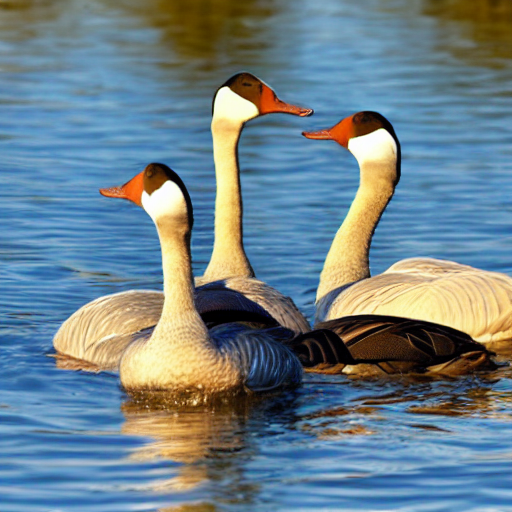}
	    & 
	\includegraphics[width=0.14\linewidth]{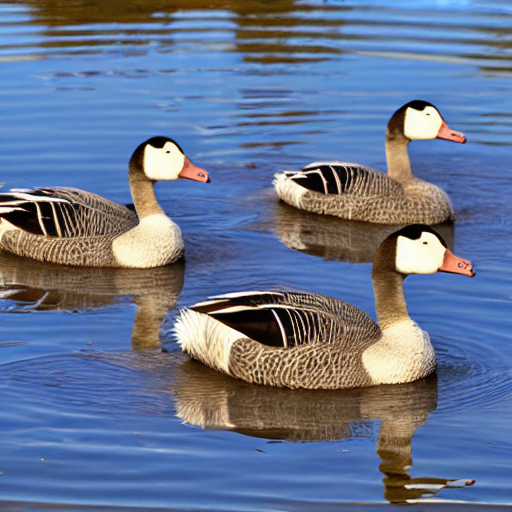}
    \tabularnewline
	\end{tabular}
\hfill{}
\par\end{centering}
\caption{Our \textbf{{\ours}} can faithfully generate multiple objects based on detailed textual description. Compared to prior state-of-the-art semantic nursing technique for text-to-image synthesis, Attend \& Excite\newcite{chefer2023attendandexcite}, our approach exhibits superior alignment with the input prompt and maintain a higher level of realism.
} 
\label{fig:teaser}
\vspace{-0.5em}
\end{figure*}

%%%%%%%%% BODY TEXT

\section{Introduction}\label{sec:intro}
In the realm of text-to-image~(T2I) synthesis, large-scale generative models\newcite{rombach2022SD,ramesh2022dalle2,saharia2022imagen,balaji2022ediffi,chang2023muse,yu2022parti,kang2023gigagan} have recently achieved significant progress and demonstrated exceptional capacity to generate stunning photorealistic images.
However, it remains challenging to synthesize images that fully comply with the given prompt input\newcite{chefer2023attendandexcite,marcus2022very,feng2023structureDiffusion,wang2022diffusiondb}. 
There are two well-known semantic issues in text-to-image synthesis, i.e., \myquote{missing objects} and \myquote{attribute binding}. \myquote{Missing objects} refers to the phenomenon that not all objects mentioned in the input text faithfully appear in the image. \myquote{Attribute binding} represents the critical compositionality problem that the attribute information, e.g., color or texture, is not properly aligned to the corresponding object or wrongly attached to the other object.
To mitigate these issues, recent work Attend \& Excite ({\AandE})\newcite{chefer2023attendandexcite} has introduced the concept of Generative Semantic Nursing (GSN). The core idea lies in updating latent codes on-the-fly such that the semantic information in the given text can be better incorporated within pretrained synthesis models. 
% one slightly shifts the latent code at each timestep of the denoising process such that the latent is encouraged to better consider the semantic information passed from the input text prompt. We propose a fo

As an initial attempt {\AandE}\newcite{chefer2023attendandexcite}, building upon the powerful open-source T2I model Stable Diffusion (SD)\newcite{rombach2022SD}, leveraged cross-attention maps for optimization.
Since cross-attention layers are the only interaction between the text prompt and the diffusion model, the attention maps %significantly 
have significant
impact on the generation process. 
%Hence, 
To enforce the object occurrence, {\AandE} defined a loss objective that attempts to maximize the maximum attention value for each object token. 
Although showing promising results on simple composition, e.g., \myquote{a cat and a frog}, we observed unsatisfying outcomes when the prompt becomes more complex, as illustrated in \cref{fig:teaser}. {\AandE} fails to faithfully synthesize the \myquote{train} or \myquote{dog} in the first two examples, and miss one \myquote{goose} in the third one. 
We attribute this to the suboptimal loss objective, which only considers the single maximum value and does not take the spatial distribution into consideration. 
As the complexity of prompts increases, token competition intensifies. 
The single excitation of one object token may overlap with others, leading to the suppression of one object by another (e.g., missing \myquote{train} in \cref{fig:teaser}) or to hybrid objects, exhibiting features of both semantic classes  (e.g., mixed dog-turtle in \cref{fig:attention_vis}).
Similar phenomenon has been observed in \newcite{tang2023daam} as well.

In this work, we propose a novel objective function for GSN. We maximize the total variation of the attention map to prompt multiple, spatially distinct attention excitations. 
By spatially distributing the attention for each token, we enable the generation of all objects mentioned in the prompt, even under high token competition.
Intuitively, this corresponds to \emph{dividing} the attention map into multiple regions.  Besides, to mitigate the attribute \emph{binding} issue, we propose a Jensen-Shannon divergence (JSD) based binding loss to explicitly align the distribution between excitation of each object and its attributes. Thus, we term our method {\ours}.
Our main contributions can be summarized as: 
(i) We propose a novel total-variation based attendance loss enabling presence of multiple objects in the generated image.
(ii) We propose a JSD-based attribute binding loss for faithfull attribute binding.
(iii) Our approach exhibits outstanding capability of generating images fully adhering to the prompt, outperforming {\AandE} on several benchmarks involving complex descriptions. 

\vspace{-0.2em}
\section{Related Work}\label{sec:related_work}
\vspace{-0.8em}

%%%%%%%%%%%%%%%%%%%%%%%%%%%%%%%%%%%%%%%%
\paragraph{Text-to-Image Synthesis.}
%%%%%%%%%%%%%%%%%%%%%%%%%%%%%%%%%%%%%%%%
With the rapid emergence of diffusion models\newcite{ho2020ddpm,song2020ddim,nichol2021improved}, recent large-scale text-to-image models such as eDiff-I\newcite{balaji2022ediffi}, Stable Diffusion\newcite{rombach2022SD}, Imagen\newcite{saharia2022imagen}, or {DALL$\cdot$E~2}\newcite{ramesh2022dalle2} have achieved impressive progress.  
Despite synthesizing high-quality images, it remains challenging to produce results that properly comply with the given text prompt.
A few recent works\newcite{feng2023structureDiffusion,chefer2023attendandexcite} aim at improving the semantic guidance purely based on the text prompt without model fine-tuning.
StructureDiffusion\newcite{feng2023structureDiffusion} used language parsers for hierarchical structure extraction, to ease the composition during generation. 
Attend \& Excite ({\AandE})\newcite{chefer2023attendandexcite} optimizes cross-attention maps during inference time by maximizing the maximum attention value of each object token to encourage object presence. 
However, we observed that {\AandE} struggles with more complex prompts. 
In contrast, our {\ours} fosters the stimulation of multiple excitations, which aids in holding the position amidst competition from other tokens. Additionally, we incorporate a novel binding loss that explicitly aligns the object with its corresponding attribute, yielding more accurate binding effect.

%%%%%%%%%%%%%%%%%%%%%%%%%%%%%%%%%%%%%%%%
\paragraph{Total Variation.}
%%%%%%%%%%%%%%%%%%%%%%%%%%%%%%%%%%%%%%%%
Total variation (TV) measures the differences between neighbors. Thus, minimization encourages smoothness that was used in different tasks, e.g., denoising\newcite{caselles2015total}, image restoration\newcite{chan2006total}, and segmentation\newcite{sun2011tvSeg}, just to name a few. Here, we use TV for a different purpose. We seek to divide attention maps into multiple excited regions. Thus, we choose TV \emph{maximization} to enlarge the amount of local changes in attention maps over the image such that diverse object regions are encouraged to emerge. As a result, we enhance the chance of generating each desired object while concurrently competing with other objects.

%%%%%%%%%%%%%%%%%%%%%
%  Method overview  %
%%%%%%%%%%%%%%%%%%%%%
\begin{figure*}[t]
\centering
\includegraphics[width=0.9\linewidth]{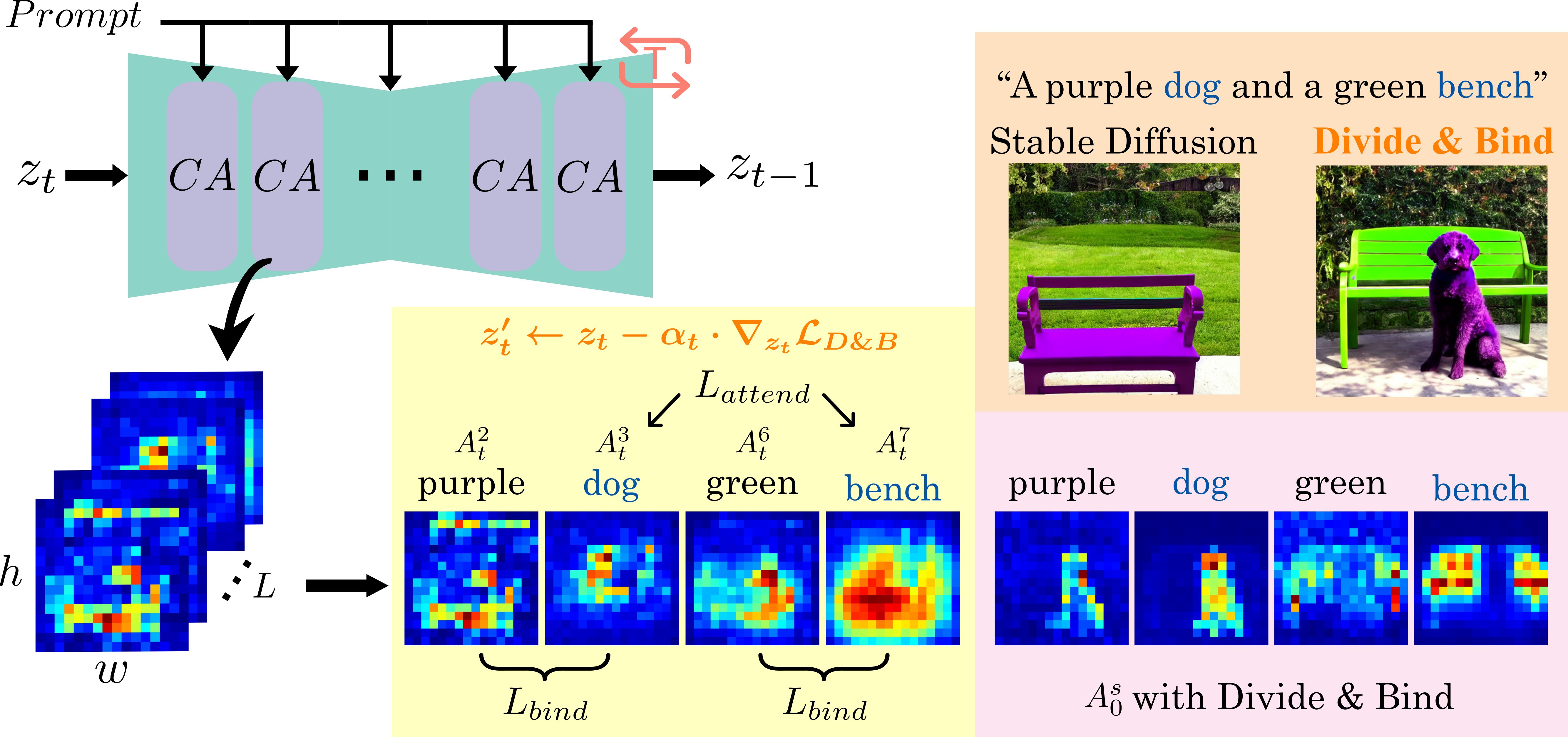}
%\vspace{-0.5em}
\caption{
Method overview. We perform latent optimization on-the-fly based on the attention maps of the object tokens with our TV-based $L_{attend}$ and JSD-based $L_{bind}$. 
}
\label{fig:overview}
\end{figure*}

%\vspace{-1em}
%%%%%%%%%%%%%%%%%%%%%%%%%%%%%%%%%%%
\section{Preliminaries}\label{sec:bg}
%%%%%%%%%%%%%%%%%%%%%%%%%%%%%%%%%%%
%\vspace{-0.7em}

%%%%%%%%%%%%%%%%%%%%%%%%%%%%%%
\paragraph{Stable Diffusion (SD).}
%%%%%%%%%%%%%%%%%%%%%%%%%%%%%%
We implement our method based on the open-source state-of-the-art T2I model SD \newcite{rombach2022SD}, which belongs to the family of latent diffusion models (LDMs). LDMs are two-stage methods, consisting of an autoencoder and a diffusion model trained in the latent space. In the first stage, the encoder $\mathcal{E}$ transforms the given image $x$ into a latent code $z = \mathcal{E}(x)$, then $z$ is mapped back to the image space by the decoder $\mathcal{D}$. The autoencoder is trained to reconstruct the given image, i.e.~$\mathcal{D}(\mathcal{E}(x)) \approx x$. In the second stage, a diffusion model\newcite{ho2020ddpm,nichol2021improved} is trained in the latent space of the autoencoder.
During training, 
we gradually add noise to the original latent $z_0$ with time, resulting in $z_t$. %
Then the UNet\newcite{ronneberger2015unet} denoiser $\epsilon_{\theta}$ is trained with a denoising objective to predict the noise $\epsilon$ that is added to $z_0$: 
\vspace{-0.5em}
\begin{align} \label{eq:ldm-loss}
\mathcal{L} = \mathbb{E}_{z\sim \mathcal{E}(x),\epsilon\sim N(0,I), c, t}  \left[\norm{\epsilon - \epsilon_{\theta} ( z_t, c, t) }^2 \right], 
\end{align}
where $c$ is the conditional information, e.g., text. During inference, given $z_T$ randomly sampled from Gaussian distribution, UNet outputs noise estimation and gradually removes it, finally producing the clean version $z_0$.

%%%%%%%%%%%%%%%%%%%%%%%%%%%%%%%%%%%%%%%%%%%%%%%%%%%%
\paragraph{Cross-Attention in Stable Diffusion.}
%%%%%%%%%%%%%%%%%%%%%%%%%%%%%%%%%%%%%%%%%%%%%%%%%%%%
In SD,  a frozen CLIP text encoder \newcite{radford2021clip} is adopted to embed the text prompt $\mathcal{P}$ into a sequential embedding as the condition $c$, which is then injected into UNet through cross-attention (CA) to synthesize text-complied images.
The CA layers take the encoded text embedding and project it into queries $Q$ and values $V$. The keys $K$ are mapped from the intermediate features of UNet. The attention maps are then computed by $A_t = \mathit{Softmax}(\frac{QK^T}{\sqrt{d}})$, where $t$ indicates the time step, Softmax is applied along the channel dimension. The attention maps $A_t$ can be reshaped into $\mathbb{R}^{h\times w \times L}$, where $h,w$ is the resolution of the feature map, $L$ is the sequence length of the text embedding. %and we omit the batch size dimension for simplicity.
Further, we denote the cross-attention map that corresponds to the $s$th text token as $A_t^s \in \mathbb{R}^{h\times w}$, see an illustration in \cref{fig:overview}. 
One known issue of SD is that not all objects are necessarily present in the final image\newcite{chefer2023attendandexcite,liu2022compositional,wang2022diffusiondb}, while, as shown in\newcite{balaji2022ediffi,hertz2022prompt2prompt}, the high activation region of the corresponding attention map strongly correlates to the appearing pixels belonging to one specific object in the final image.
Hence, the activation in the attention maps is an important signal and an influencer in the semantic guided synthesis.

%%%%%%%%%%%%%%%%%%%%%%%%%%%%%%%%%%%%%%%%%
\section{Method}\label{sec:method}
%%%%%%%%%%%%%%%%%%%%%%%%%%%%%%%%%%%%%%%%%
Given the recognized significance of the cross-attention maps in guiding semantic synthesis, our method aims at optimizing the latent code at inference time to excite them based on the text tokens. 
We employ the generative semantic nursing (GSN) method (\cref{subsec:GSN}) for latent code optimization, and propose a novel loss formulation (\cref{subsec:our_method}). It consists of two parts, i.e.~\emph{divide} and \emph{bind}, which encourages object occurrence and attribute binding respectively.

\begin{figure*}[t]
\begin{centering}
\setlength{\tabcolsep}{0.0em}
\renewcommand{\arraystretch}{0}
\par\end{centering}
\begin{centering}
%\vspace{-2em}
\hfill{}%
	\begin{tabular}{
    m{1.5em}<{\centering} %@{\hspace{1em}}
    m{0.15\linewidth}<{\centering} @{\hspace{-1em}}
    m{0.15\linewidth}<{\centering}  @{}%@{\hspace{-5em}}
    m{0.15\linewidth}<{\centering} @{}
    m{0.15\linewidth}<{\centering}
    }
    &
    \multicolumn{4}{c}{\hspace{-0.02\linewidth}\myquote{A \textblue{dog} and a \textblue{turtle} on the street, snowy scene}} 
    
    \tabularnewline
    &
    $x_0$
    & \hspace{3em}$\hat{x_0}^{(t)}$\hspace{-1em}
    & %{\hspace{4em} } %{\footnotesize{}}
    \hspace{2em}dog
    & %{\footnotesize{}} {\hspace{4em} }
     \hspace{-1em}turtle
	%\vspace{0.02cm} 
    %% Stable Diffsuion
    \tabularnewline
    \multirow{1}{*}{\rotatebox{90}{
        \hspace{2.2em}
        \begin{tabular}{c}
        \small Stable\\ \small Diffusion 
        \end{tabular}
        \hspace{-2.2em}
    }} 
    &
    \includegraphics[ height=0.09\textheight]{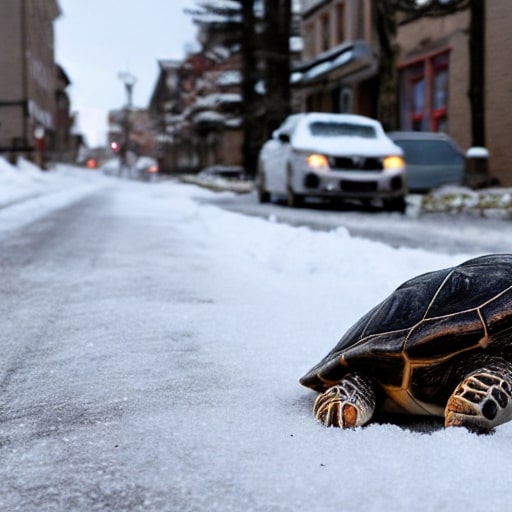}
    &
    \multicolumn{3}{c}{\begin{tabular}{c}
    %\animategraphics[autoplay,loop,height=0.09\textheight]{3}{fig/attn_gif_v2/SD/}{0}{41}
    \includegraphics[height=0.09\textheight]{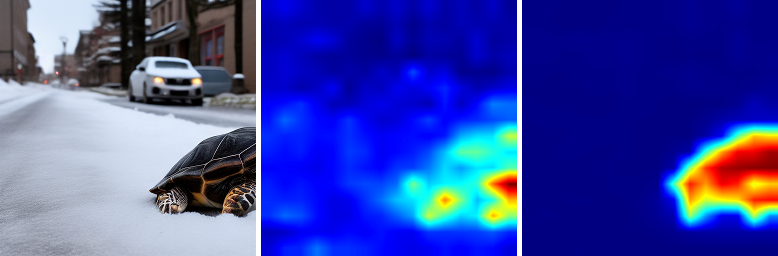}
    \end{tabular}} 
    
	 %% Attend and Excite
  \tabularnewline
  \multirow{1}{*}{\rotatebox{90}{
        \hspace{2.4em}
        \begin{tabular}{c<{\centering}}
        \small
        Attend \& \\ \small Excite
        \end{tabular}
        \hspace{-2.4em}
    }} 
    &
    \includegraphics[height=0.09\textheight]{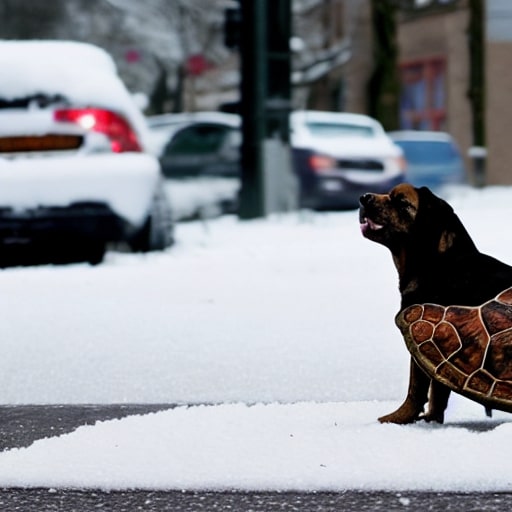}
    &
    \multicolumn{3}{c}{\begin{tabular}{c}
    %\animategraphics[autoplay,loop,height=0.09\textheight]{3}{fig/attn_gif_v2/AE/}{0}{41}
    \includegraphics[height=0.09\textheight]{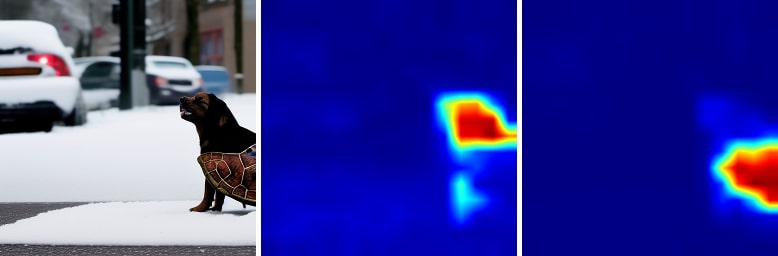}
    \end{tabular}} 
	 \tabularnewline
  %% Ours
    \multirow{1}{*}{\rotatebox{90}{
        \hspace{3.5em}
        \begin{tabular}{c}
        \small
        \textbf{{\ours}}\\
        \small (Ours) 
        \end{tabular}
        \hspace{-3.5em}
    }} 
    &
    \includegraphics[height=0.09\textheight]{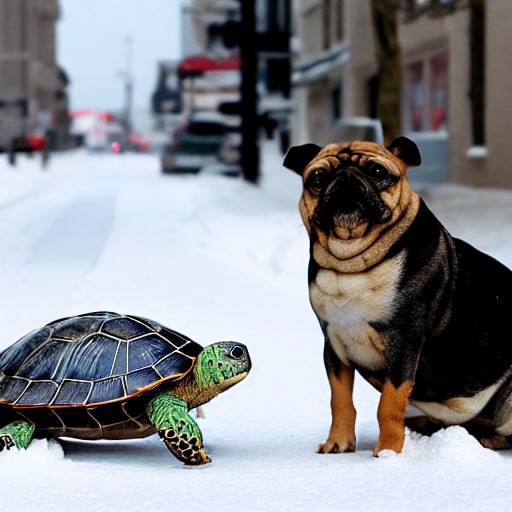}
    &
    \multicolumn{3}{c}{\begin{tabular}{c}
    %\animategraphics[autoplay,loop,height=0.09\textheight]{3}{fig/attn_gif_v2/TV/}{0}{41}
    \includegraphics[height=0.09\textheight]{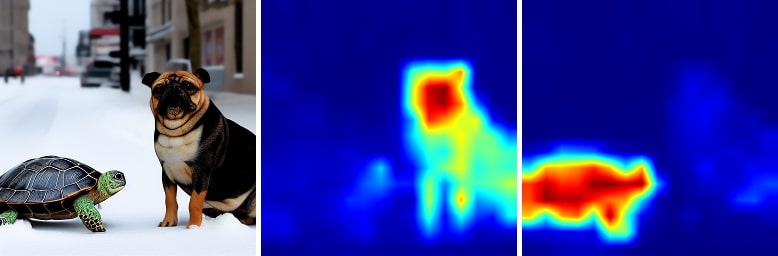}
    \end{tabular}} 

    \tabularnewline
	\end{tabular}
\hfill{}
\par\end{centering}
\caption{
Cross-attention visualization in different timesteps for each object token and predicted clean image $\hat{x_0}^{(t)}$.
Note that this is GIF, video version can be found on the \href{https://sites.google.com/view/divide-and-bind}{project page}.
} 
\label{fig:attention_vis}
%\vspace{-1em}
\end{figure*}

\begin{figure*}[t]
\begin{centering}
\setlength{\tabcolsep}{0.0em}
\renewcommand{\arraystretch}{0}
\par\end{centering}
\begin{centering}
%\vspace{-2em}
\hfill{}%
	\begin{tabular}{@{\hspace{-0.25em}}c@{\hspace{0.2em}}c@{\hspace{0.2em}}c@{\hspace{0.2em}}c@{\hspace{0.6em}}c@{\hspace{0.2em}}c@{\hspace{0.2em}}c}
	\centering
    &
    \multicolumn{3}{c}{\begin{tabular}{c}\myquote{A purple \textblue{dog} and a green \textblue{bench} \\on the street, snowy scene}\end{tabular}} 
    & 
    \multicolumn{3}{c}{\begin{tabular}{c}\myquote{A purple \textblue{crown}  and a blue \textblue{bench}}\end{tabular}} 
    \tabularnewline
	&
    & purple
	& dog 
    & 
    & purple
	& crown
	\vspace{0.02cm} 
    \tabularnewline
    \multirow{1}{*}{ \rotatebox{90}{\hspace{4.0em}  w/o \ $L_{bind}$ \hspace{-4.0em} }} 
    &
	\includegraphics[width=0.14\linewidth, height=0.08\textheight]{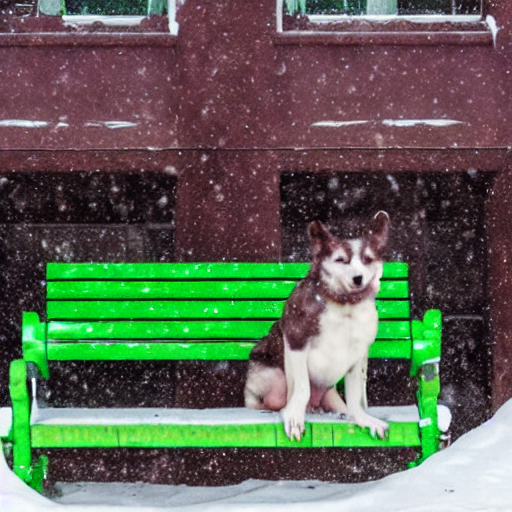}
	      &
	\includegraphics[width=0.14\linewidth, height=0.08\textheight]{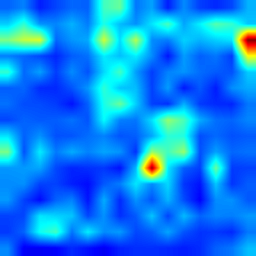}
	    & 
	\includegraphics[width=0.14\linewidth, height=0.08\textheight]{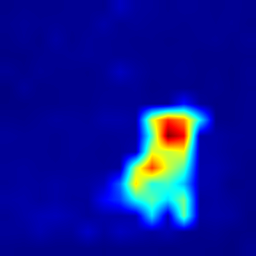}
        & 
    \includegraphics[width=0.14\linewidth, height=0.08\textheight]{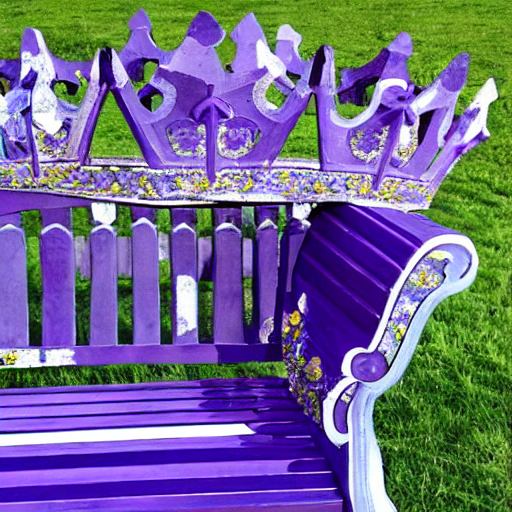}
	      & 
	\includegraphics[width=0.14\linewidth, height=0.08\textheight]{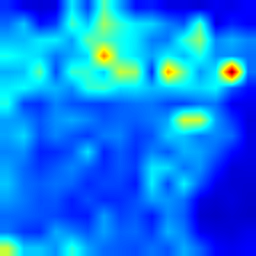}
	    & 
	\includegraphics[width=0.14\linewidth, height=0.08\textheight]{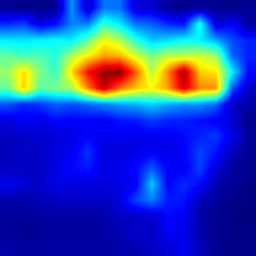}
	 %% + jSD
  \tabularnewline
  \multirow{1}{*}{ \rotatebox{90}{\hspace{3.8em}  w/- \ $L_{bind}$  \hspace{-3.8em} }} 
    &
    \includegraphics[width=0.14\linewidth, height=0.08\textheight]{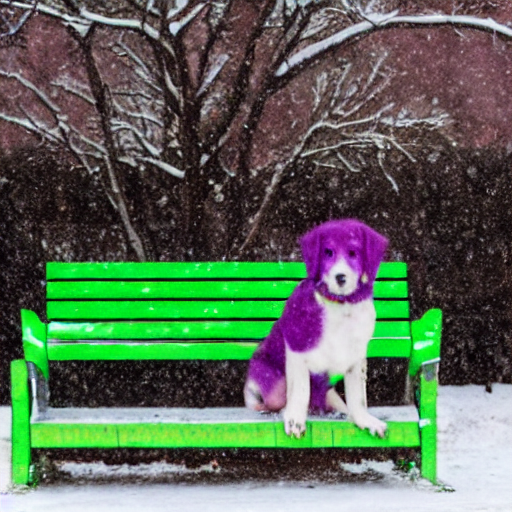}
	      &
	\includegraphics[width=0.14\linewidth, height=0.08\textheight]{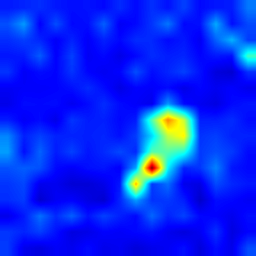}
	    & 
	\includegraphics[width=0.14\linewidth, height=0.08\textheight]{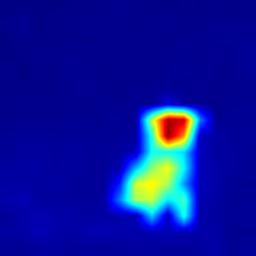}
        & 
    \includegraphics[width=0.14\linewidth, height=0.08\textheight]{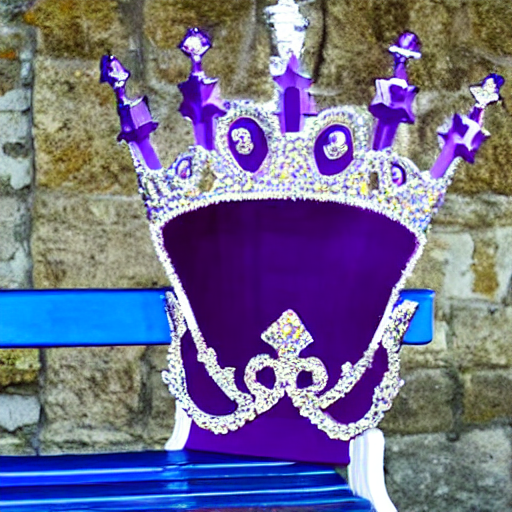}
	      & 
	\includegraphics[width=0.14\linewidth, height=0.08\textheight]{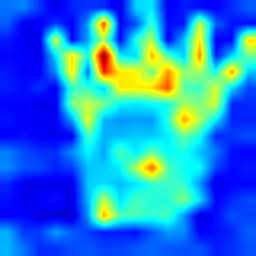}
	    & 
	\includegraphics[width=0.14\linewidth, height=0.08\textheight]{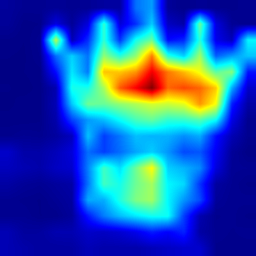}
	 \tabularnewline
	\end{tabular}
\hfill{}
\par\end{centering}
\vspace{-0.3em}
\caption{Binding loss ablation. $L_{bind}$ aligns the excitation of attribute and object attention.
} 
\label{fig:jsd_ablation}

\end{figure*}
%%%%%%%%%%%%%%%%%%%%%%%%%%%%%%%%%%%%%%%%%%%
\subsection{Generative Semantic Nursing (GSN)}\label{subsec:GSN}
%%%%%%%%%%%%%%%%%%%%%%%%%%%%%%%%%%%%%%%%%%%
To improve the semantic guidance in SD during inference, 
one pragmatic way is via latent code optimization at each time step of sampling, i.e.~GSN \newcite{chefer2023attendandexcite}
\vspace{-0.5em}
\begin{align} \label{eq:update-final}
z_t'  \leftarrow z_t - \alpha_t \cdot  \nabla_{z_t} \mathcal{L},
\end{align} % _{\mathit{obj}}
where $\alpha_t$ is the updating rate and $\mathcal{L}$ is the loss to encourage the faithfulness between the image and text description, e.g.~object attendances and attribute binding. GSN has the advantage of avoiding fine-tuning SD. 

As the text information is injected into the UNet of SD via cross attention layers, it is natural to set the loss $\mathcal{L}$ with the cross attention maps as the inputs. Given the text prompt $\mathcal{P}$ and a list of object tokens $S$, we will have a set of attention maps $\{A_t^s\}$ for $s \in S$. Ideally, if the final image contains the concept provided by the object token $s$, the corresponding cross-attention map $A_t^s$ should show strong activation. 
To achieve this, {\AandE}\newcite{chefer2023attendandexcite}  enhances the single maximum value of the attention map, i.e.~$L_{A\&E} = -\min_{s \in S}(\max_{i,j}(A_t^s[i,j]))$. However, it
% does not consider the spatial dimension and 
does not facilitate with multiple excitations, which is increasingly important when confronted with complex prompts and the need to generate multiple instances. As shown in \cref{fig:attention_vis}, a single excitation can be easily taken over by the other competitor token, leading to missing objects in the final image. Besides, it does not explicitly address the attribute binding issue.
Instead, our {\ours} promotes the allocation of attention across distinct areas, enabling the model to explore various regions for object placement. %as indicated in the text. 
Moreover, we introduce an attribute binding regularization which explicitly encourages attribute alignment.

%%%%%%%%%%%%%%%%%%%%%%%%%%%%%%%%%%%%%%%%%%%
\subsection{{\ours}}\label{subsec:our_method}
%%%%%%%%%%%%%%%%%%%%%%%%%%%%%%%%%%%%%%%%%%%
Our proposed method {\ours} consists of a novel objective for GSN
\begin{align}\label{eq:loss}
    \min_{z_t} \mathcal{L}_{D \& B} = \min_{z_t} \mathcal{L}_{attend} + \lambda \mathcal{L}_{bind} 
\end{align}
which has two parts, the attendance loss $\mathcal{L}_{attend}$ and the binding loss $\mathcal{L}_{bind}$ that respectively enforce the object attendance and attribute binding.  $\lambda$ is the weighting factor. Detailed formulation of both loss terms is presented as follows.

%%%%%%%%%%%%%%%%%%%%%%%%%%%%%%%%%%%%%%%%%%%
\paragraph{Divide for Attendance.}
%%%%%%%%%%%%%%%%%%%%%%%%%%%%%%%%%%%%%%%%%%%
The attendance loss $L_{attend}$ is to incentivize the presence of the objects, 
thus is applied to the  text tokens associated with \emph{objects} $S$, 
\begin{align} \label{eq:tv-loss}
\mathcal{L}_{attend} = -\min_{s \in S} TV(A_t^s), \  TV(A_t^s) = \sum_{i,j} \left \vert A_t^s[i+1,j] - A_t^s[i,j]\right \vert + \left \vert A_t^s[i,j+1] - A_t^s[i,j]\right \vert
\end{align}
where $A_t^s[i,j]$ denotes the attention value of the $s$-th token at the specific location $[i,j]$ and time step $t$. The loss formulation in \cref{eq:tv-loss} is based on the the finite differences approximation of the total variation (TV) $|\nabla A_t^s|$ along the spatial dimensions. It is evaluated for each object token and we take the smallest value, i.e., representing the worst case among the all object tokens. Taking the negative TV as the loss, we essentially maximize the TV for latent optimization in \cref{eq:loss}. 
% !new! 
Since TV is essentially computed as a form of summation across the spatial dimension, it encourages large activation differences across many neighboring at different spatial locations rather than a single one, thus not only having one high activation region but also many of them. 
% !new!
Such an activation pattern in the space resembles to dividing it into different regions. The model can select some of them to display the object with single or even multiple attendances. This way, conflicts between different objects that compete for the same region can be more easily resolved. 
Furthermore, from an optimization perspective, it allows the model to search among different options for converging to the final solution. The loss is applied at the initial sampling steps. 
As can be seen from the GIF in \cref{fig:attention_vis}, for the \myquote{dog} token, regions on both left and right sides are explored in the initial phase. In the end, the left side is taken over by the \myquote{turtle} but the \myquote{dog} token covers the right side. While for SD, the \myquote{dog} token has a single weak activation, and for Attend \& Excite, it only has one single high activation region on the right that is taken over by the \myquote{turtle} later.

%%%%%%%%%%%%%%%%%%%%%%%%%%%%%%%%%%%%%%%%%%%%
\paragraph{Attribute Binding Regularization.}
%%%%%%%%%%%%%%%%%%%%%%%%%%%%%%%%%%%%%%%%%%%%
In addition to the object attendance, the given attribute information, e.g.~color or material, should be appropriately attached to the corresponding object. 
We denote the attention map of the object token and its attribute token as $A^s_t$ and  $A^{r}_t$, respectively.
For attribute binding, it is desirable that $A^{r}_t$ and $A^{s}_t$ are spatially well-aligned, i.e.~high activation regions of both tokens are largely overlapped.
To this end, we introduce $\mathcal{L}_{bind}$. After proper normalization along the spatial dimension, we can view the normalized attention maps $\widetilde{A^{r}_t}$ and $\widetilde{A^{s}_t}$ as two probability mass functions whose sample space has size $h\times w$. To explicitly encourage such alignment, we can then minimize the symmetric similarity measure Jensen–Shannon divergence (JSD) between these two distributions: 
% in log space already
% KL= p*(log(p/q)), p is the target
\vspace{-0.5em}
\begin{align} \label{eq:jsd-loss}
    \mathcal{L}_{bind} = JSD\left(\widetilde{A^r_t}\| \widetilde{A^s_t}\right).
\end{align} %\vspace{-0.5em}

Specifically, we adopt the Softmax-based normalization along the spatial dimension. When performing normalization, we also observe the benefit of first aligning the value range between the two attention maps. Namely, the original attention map of the object tokens $A^{s}_t$ have higher probability values than the ones of the attribute tokens $A^{r}_t$. Therefore, we first re-scale $A^{r}_t$ to the same range as $A^{s}_t$.
% result
As illustrated in \cref{fig:jsd_ablation}, after applying $L_{bind}$, the attribute token (e.g.~\myquote{purple}) is more localized to the correct object region (e.g.~\myquote{dog} or \myquote{crown}).

\paragraph{Implementation Details.}
%%%%%%%%%%%%%%%%%%%%%%%%%%%%%%%%%%%%%%%%%%%
The token identification process can either be done manually or automatically with the aid of GPT-3\newcite{brown2020GPT} as shown in \newcite{hu2023tifa}. Taking advantage of the in-context
learning\newcite{hu2022incontext} capability of GPT-3, by providing a few in-context examples, GPT-3 can automatically extract
the desired nouns and adjectives for new input prompts.

We inherit the choice of optimization hyperparameters from the initial attempt for GSN -  Attend \& Excite ({\AandE})\newcite{chefer2023attendandexcite}. The optimization is operated on the attention map at $16 \times 16$ resolution, as they are the most semantically meaningful ones\newcite{hertz2022prompt2prompt}. 
Based on the observation that the image semantics are determined by the initial denoising steps\newcite{liew2022magicmix,kwon2023DMsemantic}, the update is only performed from $t=T$ to $t=t_{end}$, where $T=50$ and $t_{end}=25$ in all experiments. The weight of binding loss $\lambda = 1$, if the attribute information is provided. Otherwise, $\lambda = 0$, i.e., using only the attendance loss. 

\begin{figure*}[h!]
\begin{centering}
\setlength{\tabcolsep}{0.0em}
\renewcommand{\arraystretch}{0}
\par\end{centering}
\begin{centering}
%\vspace{-2em}
\hfill{}%
	\begin{tabular}{
 @{\hspace{-0.25em}}c
 @{\hspace{0.1em}}c@{\hspace{0.2em}}c
 @{\hspace{0.6em}}c@{\hspace{0.2em}}c
 @{\hspace{0.6em}}c@{\hspace{0.2em}}c
 }
	\centering
    &
    \multicolumn{2}{c}{\begin{tabular}{c}\small 
    \myquote{A \textblue{dog} and a \textblue{cat} \\ 
    \small curled up together \\ 
    \small on a couch
    }\end{tabular}} 
    & 
    \multicolumn{2}{c}{\begin{tabular}{c} \small 
    \myquote{A black \textblue{cat}  \\ 
    \small and a red \textblue{suitcase} \\ 
    \small in the library
    }\end{tabular}} 
    &
    \multicolumn{2}{c}{\begin{tabular}{c} \small 
    \myquote{Three \textblue{sheep} \\ 
    \small standing in the field
    }\end{tabular}} 
	%\vspace{0.02cm} 
    %% Stable Diffsuion
    \tabularnewline
    \multirow{1}{*}{\rotatebox{90}{
        \hspace{4.5em}
        \begin{tabular}{c}
        \small Stable\\ \small Diffusion 
        \end{tabular}
        \hspace{-4.5em}
    }} 
        &
	\includegraphics[width=0.14\linewidth, height=0.08\textheight]{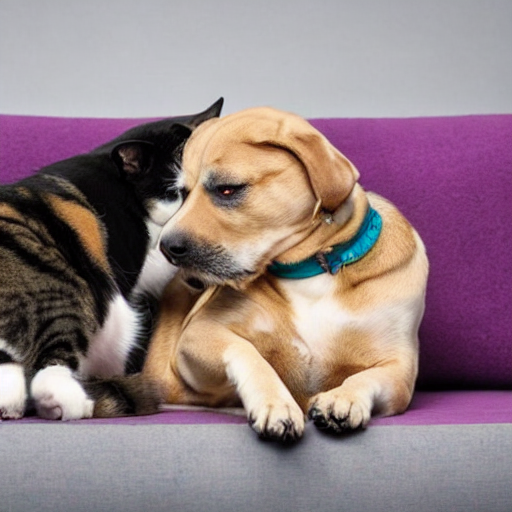}
	      &
	\includegraphics[width=0.14\linewidth, height=0.08\textheight]{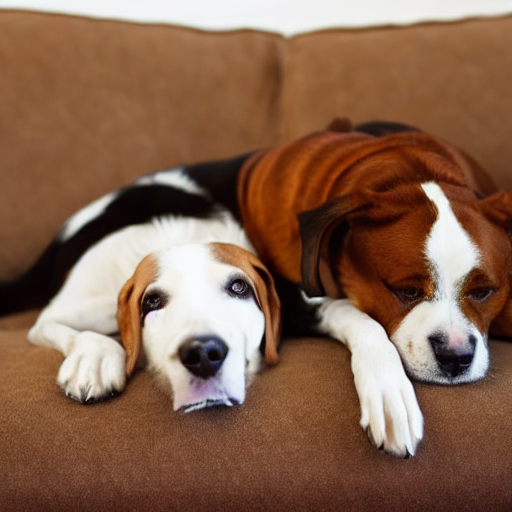}
	    & 
	\includegraphics[width=0.14\linewidth, height=0.08\textheight]{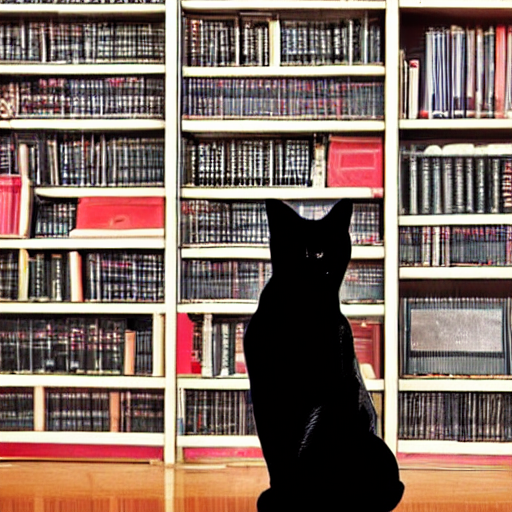}
        & 
    \includegraphics[width=0.14\linewidth, height=0.08\textheight]{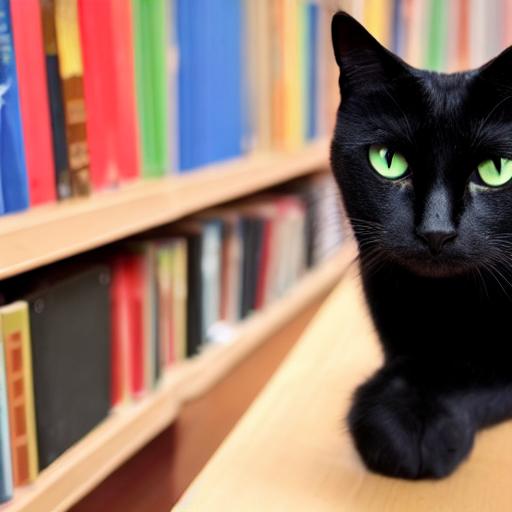}
	      & 
	\includegraphics[width=0.14\linewidth, height=0.08\textheight]{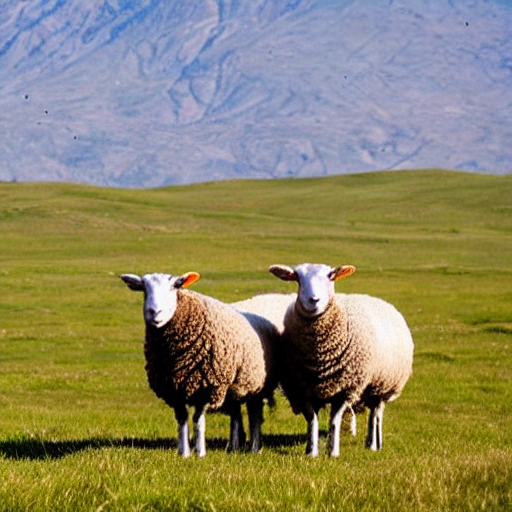}
	    & 
	\includegraphics[width=0.14\linewidth, height=0.08\textheight]{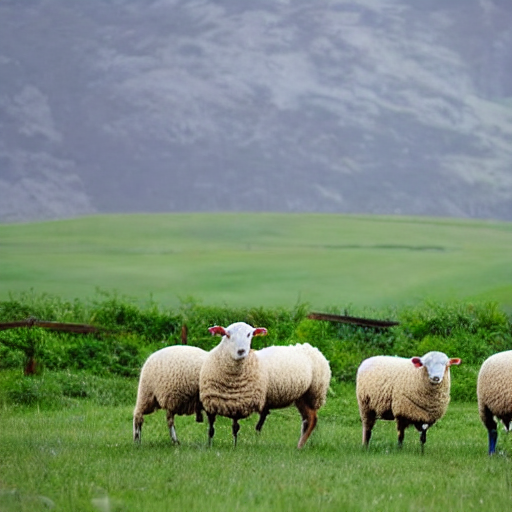}
	 %% Attend and Excite
  \tabularnewline
  \multirow{1}{*}{\rotatebox{90}{
        \hspace{4.5em}
        \begin{tabular}{c}
        \small Attend \& \\ \small Excite
        \end{tabular}
        \hspace{-4.5em}
    }} 
    &
    \includegraphics[width=0.14\linewidth, height=0.08\textheight]{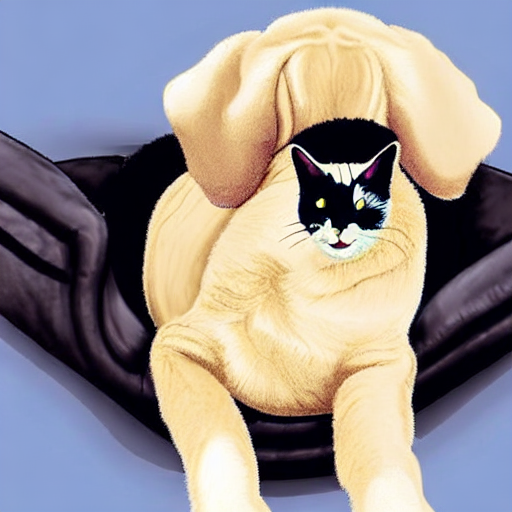}
	      &
	\includegraphics[width=0.14\linewidth, height=0.08\textheight]{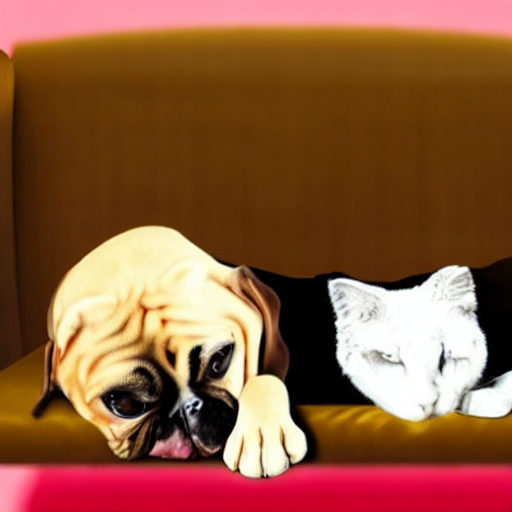}
	    & 
	\includegraphics[width=0.14\linewidth, height=0.08\textheight]{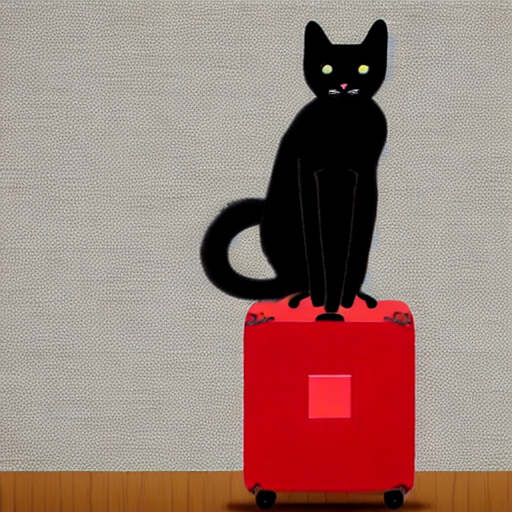}
        & 
    \includegraphics[width=0.14\linewidth, height=0.08\textheight]{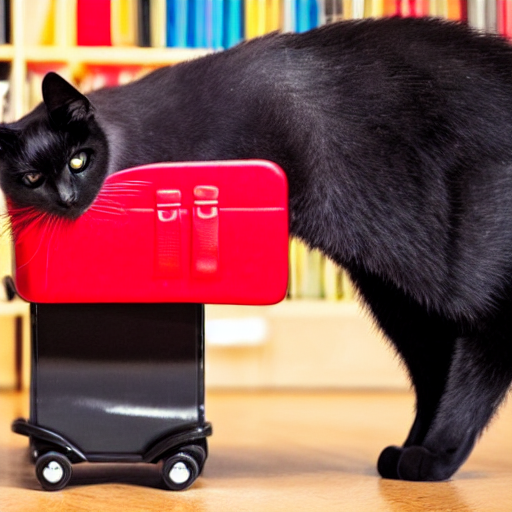}
	      & 
	\includegraphics[width=0.14\linewidth, height=0.08\textheight]{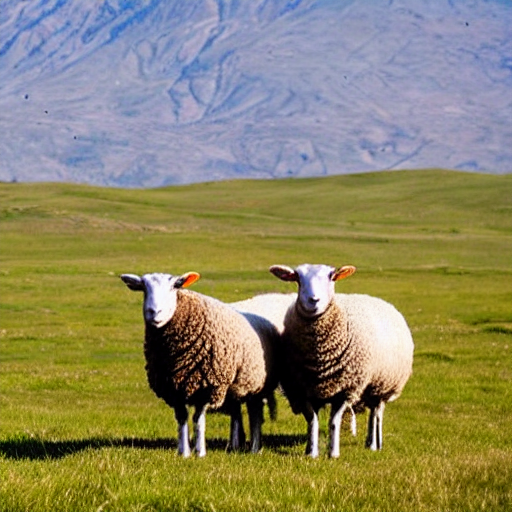}
	    & 
	\includegraphics[width=0.14\linewidth, height=0.08\textheight]{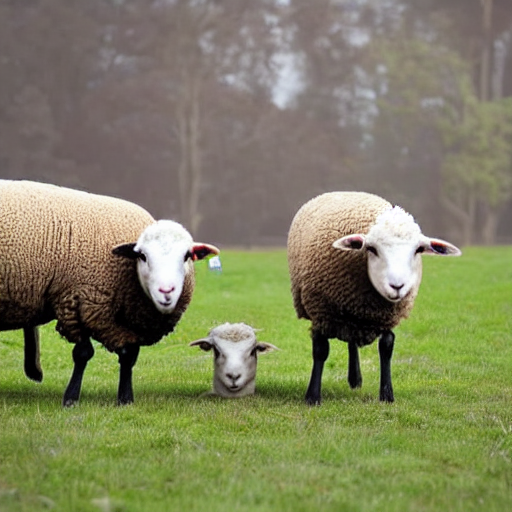}
	 \tabularnewline
  %% Ours
    \multirow{1}{*}{\rotatebox{90}{
        \hspace{4.5em}
        \begin{tabular}{c}
           \small \textbf{Divide \&} \\ \small \textbf{Bind} 
        \end{tabular}
        \hspace{-4.5em}
    }} 
    &
	\includegraphics[width=0.14\linewidth, height=0.08\textheight]{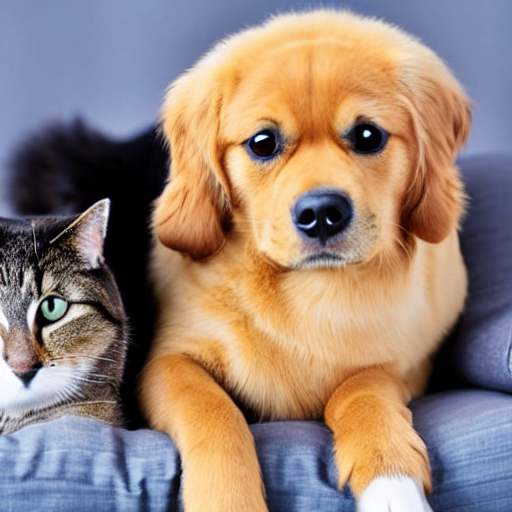}
	      &
	\includegraphics[width=0.14\linewidth, height=0.08\textheight]{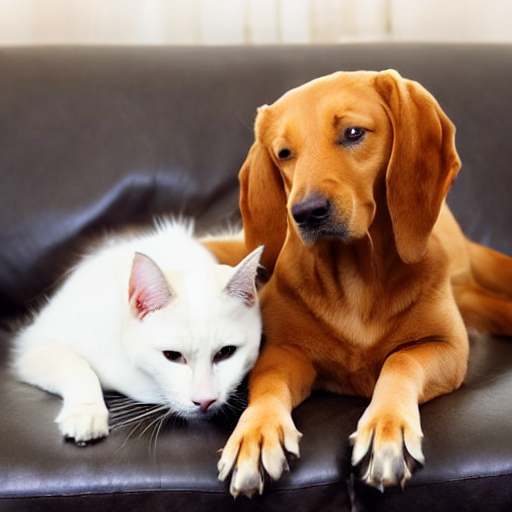}
	    & 
	\includegraphics[width=0.14\linewidth, height=0.08\textheight]{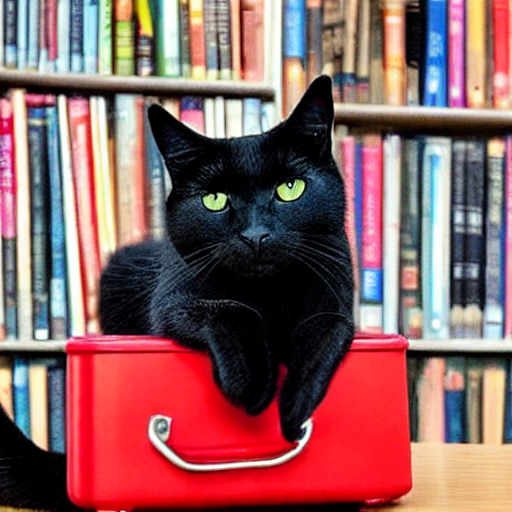}
        & 
    \includegraphics[width=0.14\linewidth, height=0.08\textheight]{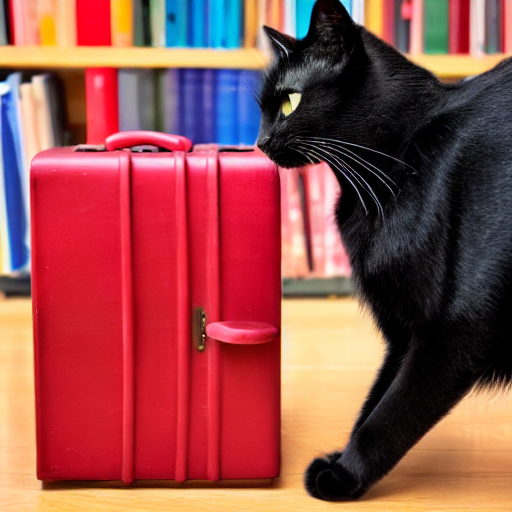}
	      & 
	\includegraphics[width=0.14\linewidth, height=0.08\textheight]{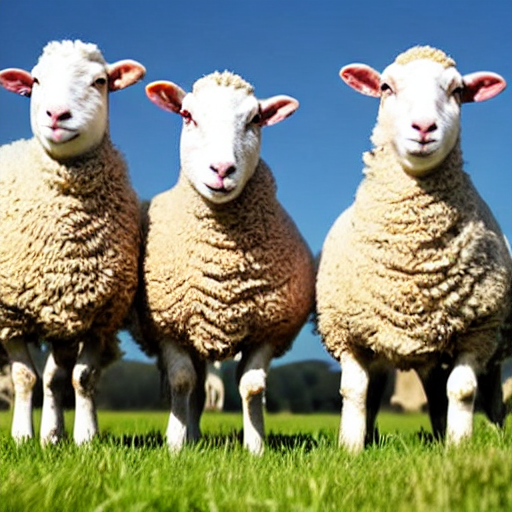}
	    & 
	\includegraphics[width=0.14\linewidth, height=0.08\textheight]{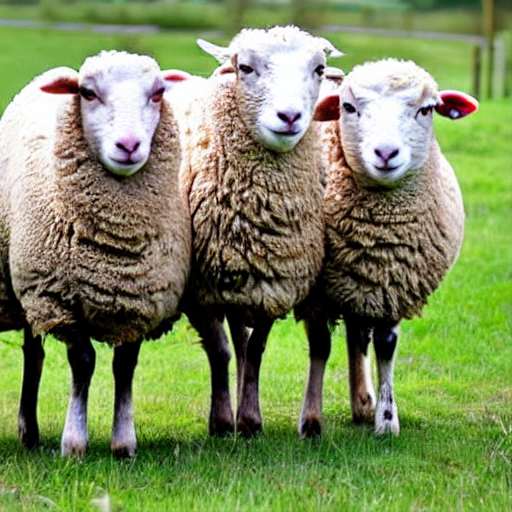}
    
   \vspace{0.5em}
    \tabularnewline
    
    %%%%%%%%%%%%%%%%%%%%%%%%%%%%%%
    % Another row of comparison %
    %%%%%%%%%%%%%%%%%%%%%%%%%%%%%%
    &
    \multicolumn{2}{c}{\begin{tabular}{c}\small 
    \myquote{A \textblue{bird} and a \textblue{bear} \\ 
    \small on the street,\\ 
    \small snowy scene
    }\end{tabular}} 
    & 
    \multicolumn{2}{c}{\begin{tabular}{c}\small  
    \myquote{A green \textblue{backpack}  \\ 
    \small and a pink \textblue{chair} \\
    \small  in the kitchen
    }\end{tabular}} 
    &
    \multicolumn{2}{c}{\begin{tabular}{c} \small 
    \myquote{One \textblue{cat} \\ 
    \small and two \textblue{dogs} 
    }\end{tabular}}  
    %% Stable Diffsuion
    \tabularnewline
    \multirow{1}{*}{\rotatebox{90}{
        \hspace{4.5em}
        \begin{tabular}{c}
        \small Stable\\ \small Diffusion 
        \end{tabular}
        \hspace{-4.5em}
    }} 
        &
	\includegraphics[width=0.14\linewidth, height=0.08\textheight]{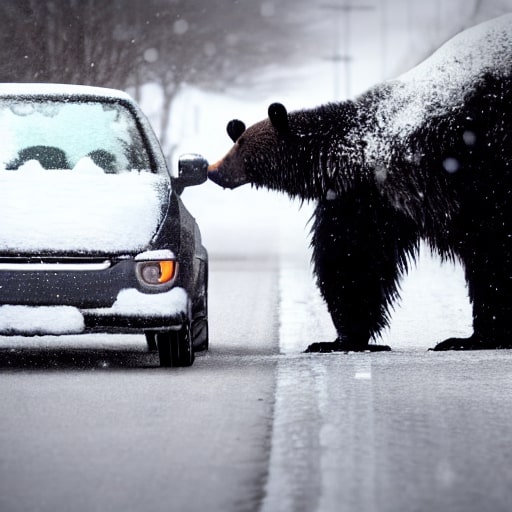}
	      &
	\includegraphics[width=0.14\linewidth, height=0.08\textheight]{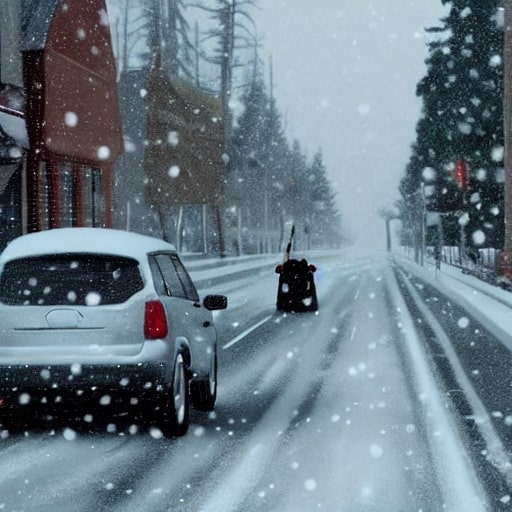}
	    & 
	\includegraphics[width=0.14\linewidth, height=0.08\textheight]{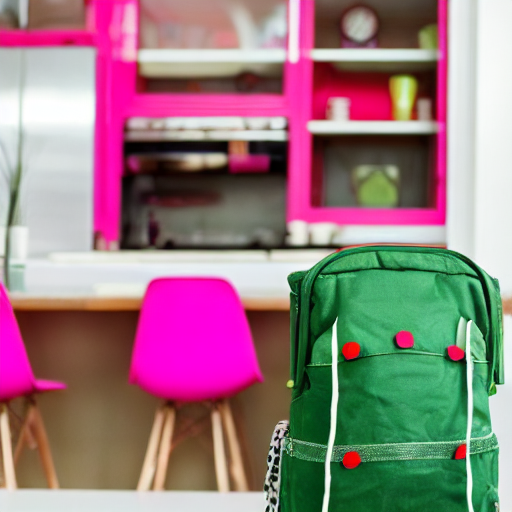}
        & 
    \includegraphics[width=0.14\linewidth, height=0.08\textheight]{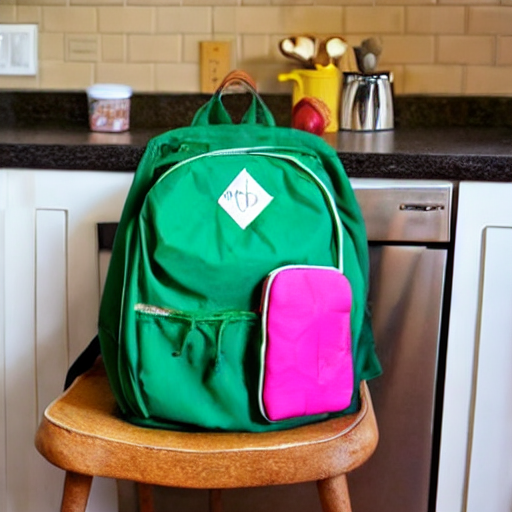}
	      & 
	\includegraphics[width=0.14\linewidth, height=0.08\textheight]{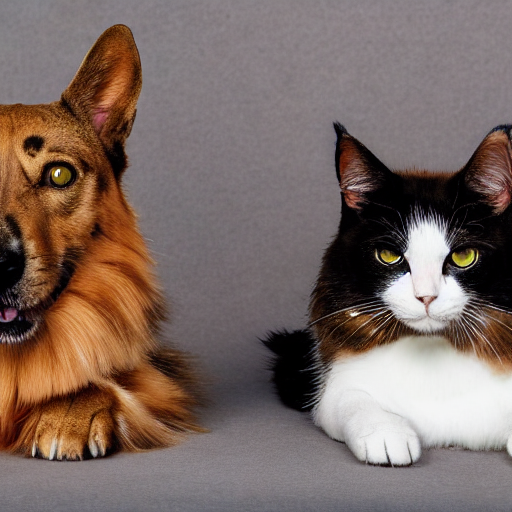}
	    & 
	\includegraphics[width=0.14\linewidth, height=0.08\textheight]{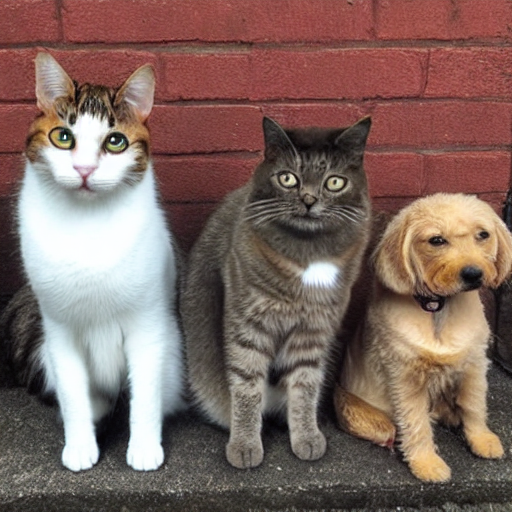}
	 %% Attend and Excite
  \tabularnewline
  \multirow{1}{*}{\rotatebox{90}{
        \hspace{4.5em}
        \begin{tabular}{c}
        \small Attend \& \\ \small Excite
        \end{tabular}
        \hspace{-4.5em}
    }} 
    &
    \includegraphics[width=0.14\linewidth, height=0.08\textheight]{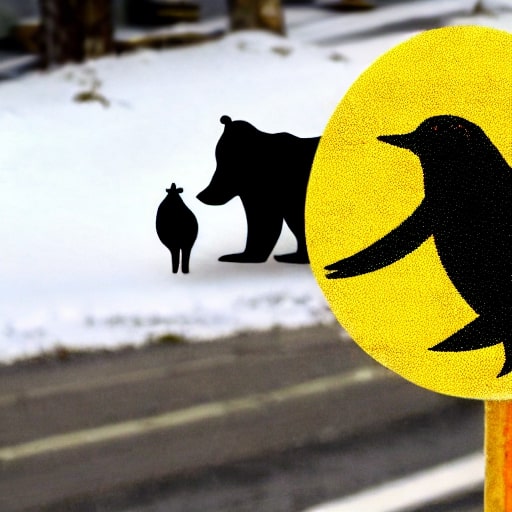}
	      &
	\includegraphics[width=0.14\linewidth, height=0.08\textheight]{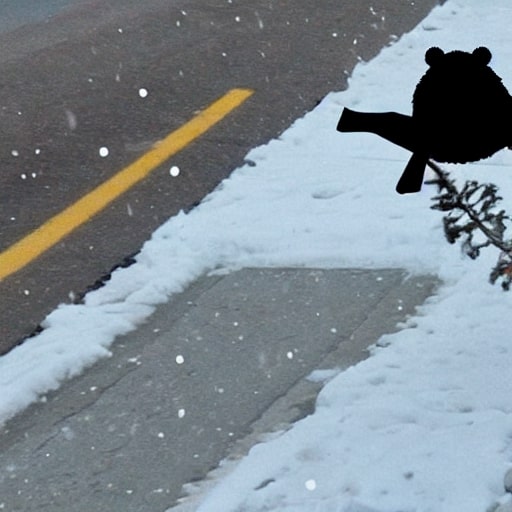}
	    & 
	\includegraphics[width=0.14\linewidth, height=0.08\textheight]{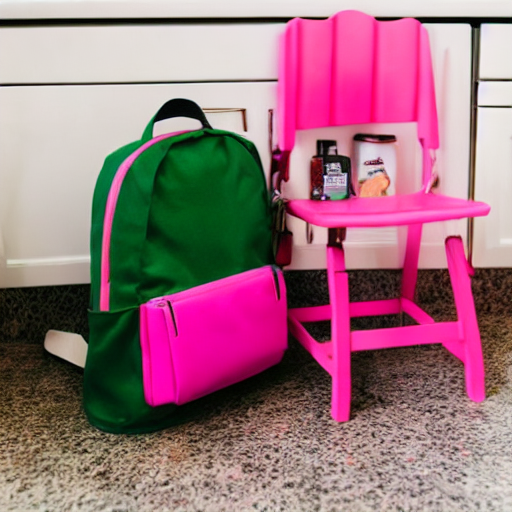}
        & 
    \includegraphics[width=0.14\linewidth, height=0.08\textheight]{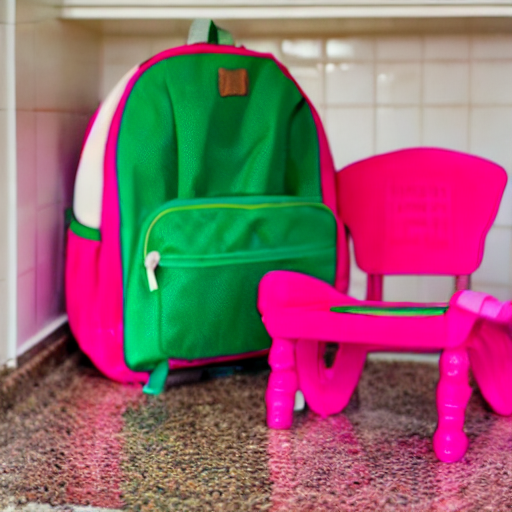}
	      & 
	\includegraphics[width=0.14\linewidth, height=0.08\textheight]{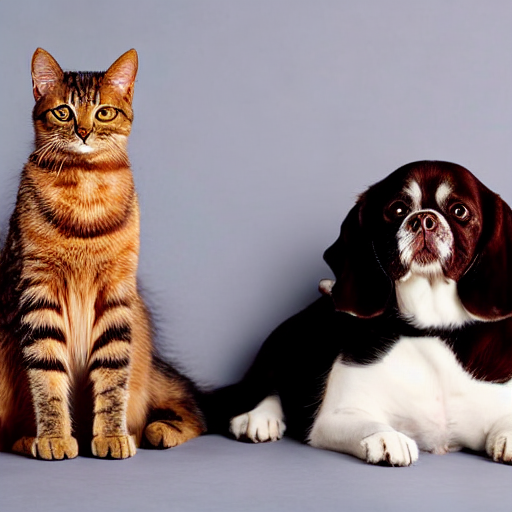}
	    & 
	\includegraphics[width=0.14\linewidth, height=0.08\textheight]{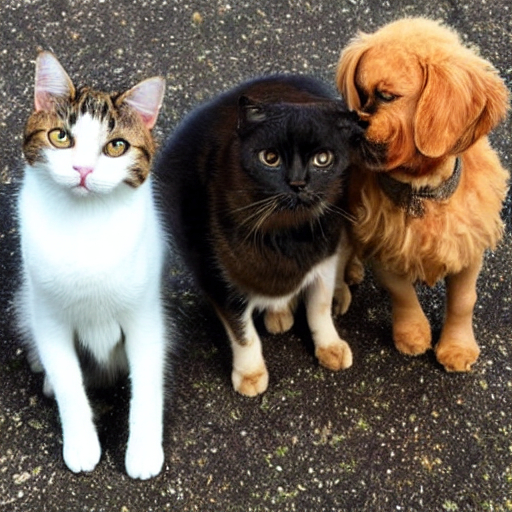}
	 \tabularnewline
  %% Ours
    \multirow{1}{*}{\rotatebox{90}{
        \hspace{4.5em}
        \begin{tabular}{c}
         \small \textbf{Divide \&} \\ \small \textbf{Bind} 
        \end{tabular}
        \hspace{-4.5em}
    }} 
    &
	\includegraphics[width=0.14\linewidth, height=0.08\textheight]{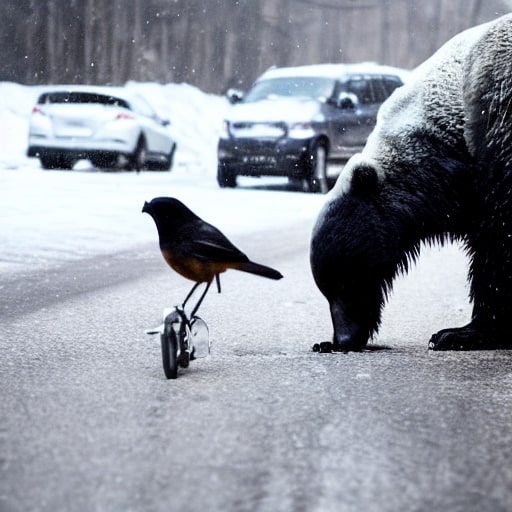}
	      &
	\includegraphics[width=0.14\linewidth, height=0.08\textheight]{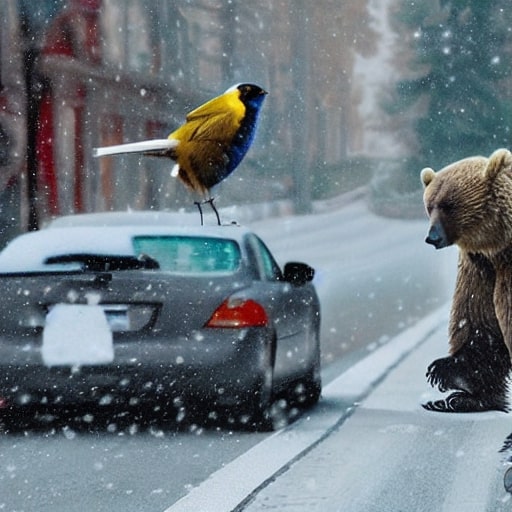}
	    & 
	\includegraphics[width=0.14\linewidth, height=0.08\textheight]{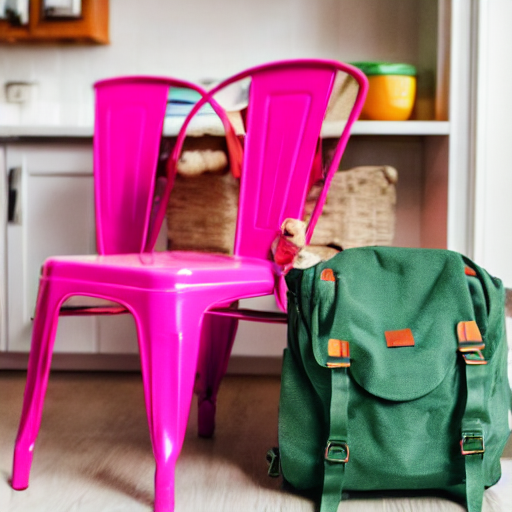}
        & 
    \includegraphics[width=0.14\linewidth, height=0.08\textheight]{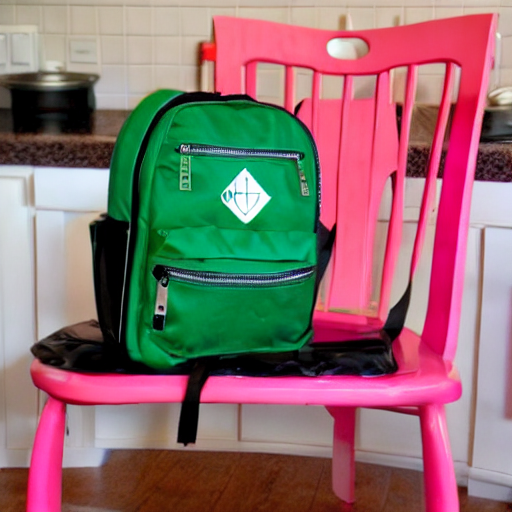}
	      & 
	\includegraphics[width=0.14\linewidth, height=0.08\textheight]{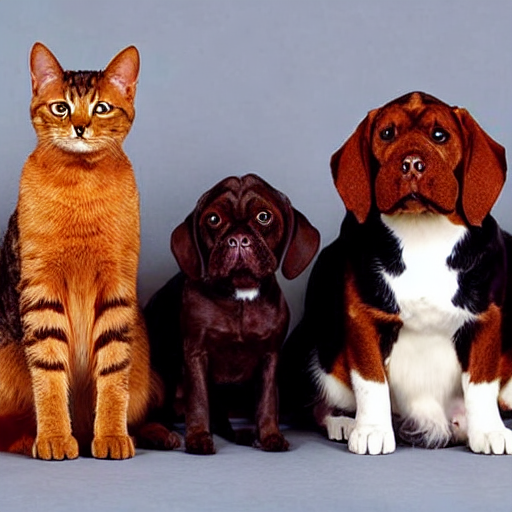}
	    & 
	\includegraphics[width=0.14\linewidth, height=0.08\textheight]{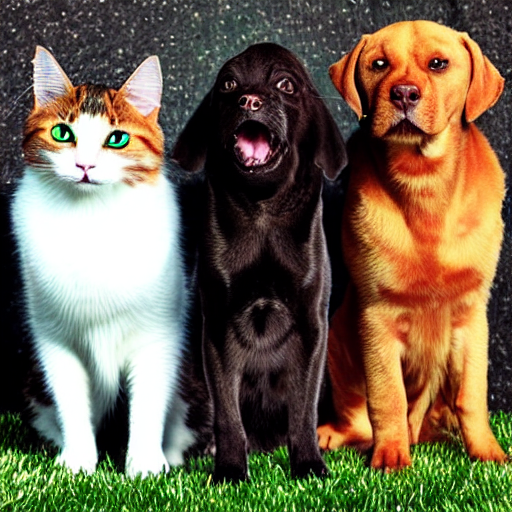}
    \tabularnewline
	\end{tabular}
\hfill{}
\par\end{centering}
\caption{Qualitative comparison in different settings with the same random seeds. Tokens used for optimization are highlighted in \textblue{blue}. Compared to others, {\ours} shows superior alignment with the input prompt while maintaining a high level of realism.
} 
\label{fig:visual_compare}
%\vspace{-0.2em}
\end{figure*}
%between Stable Diffusion, Attend and TV
\begin{table}[t]
\setlength{\tabcolsep}{0.7em}
\renewcommand{\arraystretch}{1.3}
\begin{center}
{\footnotesize %\small\footnotesize 
 %\vspace{-1.4em}
    \begin{tabular}{@{}l c c c @{}}
    %\toprule
    Evaluation Set  & Description &  Example & \# Prompt  \\ \hline
    Animal-Animal  
        &  a [animalA] and a [animalB]  
        & ``a cat and a frog''
        & 66 \\ \hline
    Color-Object   
        & \makecell[cc]{ a [colorA] [subjectA] \\ and a [colorB] [subjectB] } 
        & \makecell[cc]{ ``a green backpack \\ and  a pink chair''} 
        & 66 \\ \hline
    {\animalScene} 
        &   \makecell[cc]{a [animalA] and \\ a [animalB] [scene]}
        & \makecell[cc]{``a bird and a bear \\ in the kitchen''}
        & 56 \\ \hline
    {\colorScene}  
        &   \makecell[cc]{a [colorA] [subjectA] and \\ a [colorB] [subjectB] [scene]}
        & \makecell[cc]{``a black cat and a red suitcase \\ in the library''}
        & 60 \\ \hline
    {\multiobject} 
        & \makecell[cc]{more than two instances in the image}
        & \makecell[cc]{ ``two cats and two dogs''\\ ``three sheep standing in the field''}
        & 30 \\ \hline
    {\tifaCOCO}    
        &  \makecell[cc]{filtered COCO captions containing \\ subject related questions from TIFA}
        &  \makecell[cc]{``a dog and a cat curled up \\ together on a couch''}
        &  30 \\ \hline
    {\tifaAttr}   
        &  \makecell[cc]{filtered COCO captions containing \\ attribute related questions from TIFA}
        &  \makecell[cc]{``a red sports car is parked \\beside a black horse''}
        & 30 %\\ %\hline
    \end{tabular}
}
\end{center} 
\vspace{-0.5em}
\caption{
Description of benchmarks used for the experimental evaluation.
}
\label{tab:benchmarks}
\end{table}

\vspace{-1em}
\section{Experiments}\label{sec:experiment}

%\vspace{-0.5em}
%%%%%%%%%%%%%%%%%%%%%%%%%%%%%%%%%%%%
\subsection{Experimental Setup}
%%%%%%%%%%%%%%%%%%%%%%%%%%%%%%%%%%%%
%\vspace{-0.7em}
\paragraph{Benchmarks.}
We conduct exhaustive evaluation on seven prompt sets as summarized in \cref{tab:benchmarks}.
Animal-Animal and Color-Object are proposed in\newcite{chefer2023attendandexcite}, which simply compose two subjects and alternatively assign a color to the subject. Building on top of this, we append a postfix describing the scene or scenario to challenge the methods with higher prompt complexity, termed as {\animalScene} and {\colorScene}. Further, we introduce {\multiobject} which aims to produce multiple entities in the image. Note that different entities could belong to the same category. For instance, \myquote{one cat and two dogs} contains in total three entities and two of them are dogs. Besides the designed templates, we also filtered the COCO captions used in the TIFA benchmark\newcite{hu2023tifa} and categorize them into {\tifaCOCO} and {\tifaAttr}. There are up to four objects without any attribute assigned in {\tifaCOCO} and two objects with attributes {\tifaAttr}, respectively. Note that the attributes in {\tifaAttr} contain not only color, but also texture information, such as \myquote{a wooden bench}.

\paragraph{Evaluation metrics.}
To quantitatively evaluate the performance of our method, we used the text-text similarity from~\cite{chefer2023attendandexcite} and the recently introduced TIFA score~\cite{hu2023tifa}, which is more accurate than CLIPScore~\cite{radford2021clip} and has much better alignment with human judgment on text-to-image synthesis.
% Text-Text similarity
To compute the text-text similarity, we employ the off-the-shelf image captioning model BLIP\newcite{li2022blip} to generate captions on synthesized images. We then measure the CLIP similarity between the original prompt and all captions. 
% TIFA Score
Evaluation of the TIFA metric is based on a performance of the visual-question-answering (VQA) system, e.g.~mPLUG\newcite{li-etal-2022-mplug}. By definition, the TIFA score is essentially the VQA accuracy.
More detailed description of the TIFA evaluation protocol and evaluation on the full prompt text-image similarity and minimum object similarity from\newcite{chefer2023attendandexcite} can be found in the supp.~material.

%\vspace{-0.2em}
%%%%%%%%%%%%%%%%%%%%%%%%%%%%%%%%%%%%%%%%%%%
\subsection{Main Results}
%%%%%%%%%%%%%%%%%%%%%%%%%%%%%%%%%%%%%%%%%%%
%\vspace{-0.4em}

% Barplot comparison 

\begin{figure*}[t!]
\centering
\includegraphics[width=0.97
\linewidth]{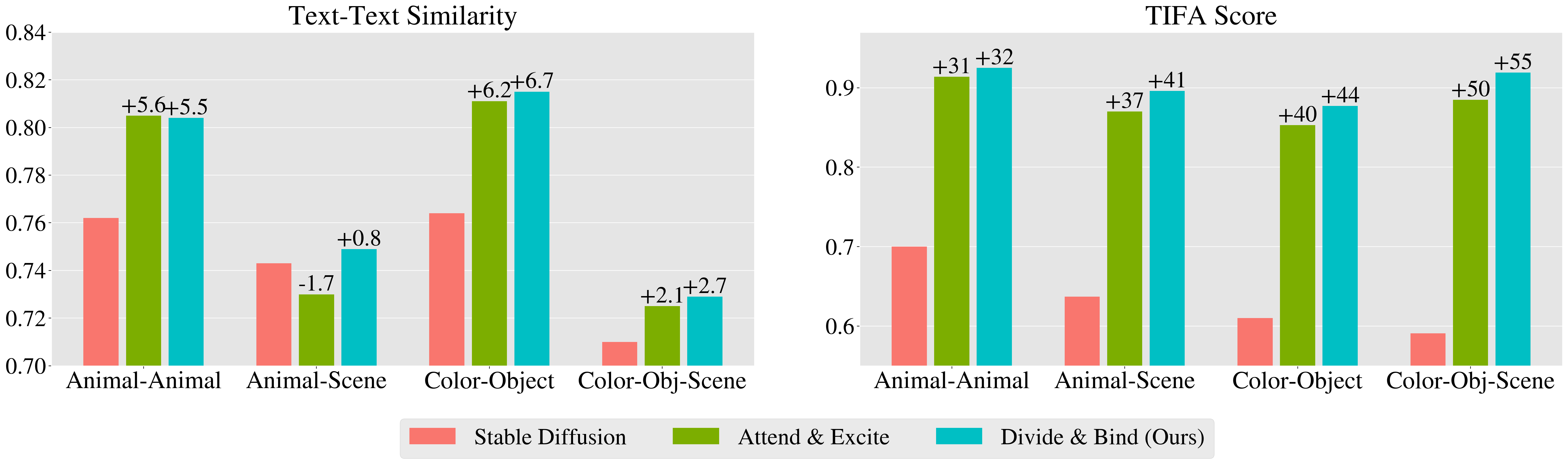}
%\vspace{-0.5em}
\caption{Quantitative comparison using Text-Text similarity and TIFA Score. {\ours} achieves comparable performance to {\AandE} on the simple Animal-Animal and Color-Object, and shows superior results on more complex text descriptions, i.e., {\animalScene} and {\colorScene}. Improvements over SD in $\%$ are reported on top of the bars.
}\label{fig:barplot}

\end{figure*}
\begin{table}[t]
\setlength{\tabcolsep}{0.5em}
\renewcommand{\arraystretch}{1.2}
\begin{center}
{\small %\small\footnotesize 
 %\vspace{-1.4em}
%\begin{minipage}{.4\linewidth}
% text-text and tifa
    \begin{tabular}{@{\extracolsep{6pt}}l  c c @{\hspace{1em}} c c @{\hspace{1em}} c  c @{}}
    %\toprule
    \multirow{2}{*}{Method}  &
    \multicolumn{2}{c}{{\multiobject}} &
    \multicolumn{2}{c}{{\tifaCOCO}}  &  \multicolumn{2}{c}{{\tifaAttr}}  \\  
    \cline{2-3} \cline{4-5} \cline{6-7}
    & Text-Text & TIFA & Text-Text & TIFA & Text-Text & TIFA \\
    \hline
    Stable Diffusion 
        & 0.786 & 0.647 & 0.823 & 0.791 & 0.790 & 0.752  \\
    Attend \& Excite  
        & \textbf{0.809} & 0.755 & 0.818 & 0.824 & 0.793 & 0.798  \\
    {\ours}
        & 0.805 & \textbf{0.785}  & \textbf{0.824} & \textbf{0.840} & \textbf{0.799} & \textbf{0.805} \\
    \end{tabular}
}
\end{center} 
\vspace{-0.4em}
\caption{
    Quantitative comparison on complex COCO-captions and {Multi-Object} generation. {\ours} 
    surpasses the other methods when it comes to handling complex prompts. 
}
\label{tab:eval_results}

\end{table}

%%%%%%%%%%%%%%%%%%%%%%%%%%%%%%%%%%%%%%
% \paragraph{A&E Setting} + scene
%%%%%%%%%%%%%%%%%%%%%%%%%%%%%%%%%%%%%%
As shown in \cref{fig:barplot}, we first quantitatively compare {\ours} with Stable Diffusion (SD)\newcite{rombach2022SD} and Attend \& Excite ({\AandE})\newcite{chefer2023attendandexcite} on Animal-Animal and Color-Object, originally proposed in \newcite{chefer2023attendandexcite}, as well as our new benchmarks {\animalScene} and {\colorScene}, which include scene description and has higher prompt complexity. 
It can be seen that {\ours} is on-par with {\AandE} on Animal-Animal and achieves  slight improvement on Color-Object. Due to the simplicity of the template, the potential of our method cannot be fully unleashed in those settings. In more complex prompts: {\animalScene} and {\colorScene}, {\ours} outperforms the other methods more evidently, especially on the TIFA score (e.g., 5\% improvement over {\AandE} in {\colorScene}). Qualitatively, both SD and {\AandE} may neglect the objects, as shown in the \myquote{bird and a bear on the street, snowy scene} example in \cref{fig:visual_compare}. Despite the absence of objects in the synthesized images, we found SD can properly generate the scene, while {\AandE} tends to ignore it occasionally, e.g.~the \myquote{library} and \myquote{kitchen} information in the second column of \cref{fig:visual_compare}). In the \myquote{a green backpack and a pink chair in the kitchen} example, both SD and {\AandE} struggle to bind the pink color with the chair only.
In contrast, {\ours}, enabled by the {\jsd} loss, demonstrates a more accurate binding effect and has less leakage to other objects or background. We provide ablation on the {\jsd} loss in the supp.~material.

%%%%%%%%%%%%%%%%%%%%%%%%%%%%%%%%%%%%%%
% \paragraph{Multi-object} 
%%%%%%%%%%%%%%%%%%%%%%%%%%%%%%%%%%%%%%
% firstly describe multi-object
Next, we evaluate the methods on {\multiobject}, where multiple entities should be generated. Visual comparison is presented in the third column of \cref{fig:visual_compare}. In the \myquote{three sheep standing in the field} example, both SD and {\AandE} only synthesize two realistic looking sheep, while the image generated by {\ours} fully complies with the prompt. For the \myquote{one cat and two dogs} example, SD and {\AandE} either miss one entity or generate the wrong species. We observe that often the result of {\AandE} resembles the one of SD. This is not surprising, as {\AandE} does not encourage attention activation in multiple regions. As long as one instance of the corresponding object token appears, the loss of {\AandE} would be low, leading to minor update.  
We also provide the quantitative evaluation in \cref{tab:eval_results}. Our {\ours} outperforms other methods by a large margin on the TIFA score, but only slightly underperforms {\AandE} on Text-Text similarity. We hypothesize that this is due to the incompetence of CLIP on counting\newcite{paiss2023teaching}, thus leading to inaccurate evaluation, as pointed out in\newcite{hu2023tifa} as well.

%%%%%%%%%%%%%%%%%%%%%%%%%%%%%%%%%%%%%%
% \paragraph{COCO} 
%%%%%%%%%%%%%%%%%%%%%%%%%%%%%%%%%%%%%%
% then  describe COCO
We also benchmark on real image captions, i.e.~{\tifaCOCO} and {\tifaAttr}, where the text structure can be more complex than fixed templates. Quantitative evaluation is provided in \cref{tab:eval_results}, where {\ours} showcases its advantages on both benchmarks over SD and {\AandE}. A visual example \myquote{a dog and a cat curled up together on a couch} is shown in \cref{fig:visual_compare}. Consistent with the observation above: while {\AandE} encourages the object occurrence, it may generate unnatural looking images. While SD, may neglect the object, its results are more realistic. {\ours} performs well with respect to both perspectives.

%%%%%%%%%%%%%%%%%%%%%%%%%%%%%%%%%%%%%%%%%%%
\paragraph{Limitations.}
%%%%%%%%%%%%%%%%%%%%%%%%%%%%%%%%%%%%%%%%%%%
\begin{figure*}[t]
\begin{centering}
\setlength{\tabcolsep}{0.0em}
\renewcommand{\arraystretch}{0}
\par\end{centering}
\begin{centering}
\hfill{}%
	\begin{tabular}{
 @{}
 c@{\hspace{0.4em}}c@{\hspace{0.4em}}c
 @{\hspace{2em}}
 c@{\hspace{0.4em}}c@{\hspace{0.4em}}c
 }
	%\centering
    %\multicolumn{3}{c}{\myquote{Two  \textblue{bananas} and two \textblue{oranges} }}
    \multicolumn{3}{c}{\myquote{A pink  \textblue{chair} and a gray \textblue{apple} }}
    & 
    \multicolumn{3}{c}{\begin{tabular}{c}\myquote{One  \textblue{dog} and three \textblue{cats}  }\end{tabular}} 
    
	%\vspace{0.02cm} 
    \tabularnewline
    % SD
	\includegraphics[width=0.14\linewidth, height=0.08\textheight]{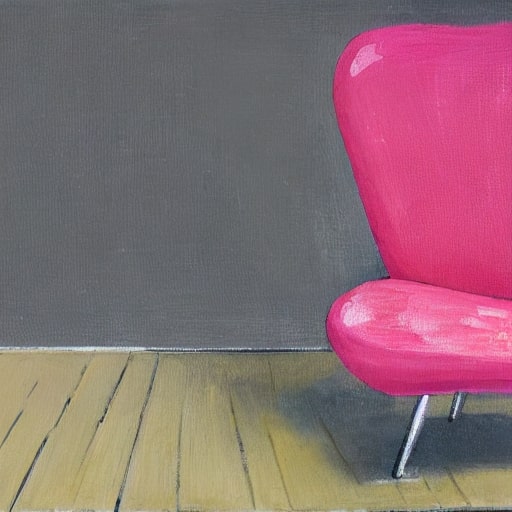}
	      &
    % AE
	\includegraphics[width=0.14\linewidth, height=0.08\textheight]{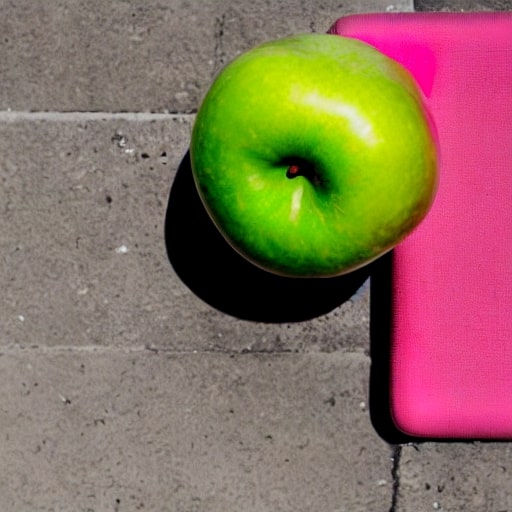}
	    & 
     % TV
	\includegraphics[width=0.14\linewidth, height=0.08\textheight]{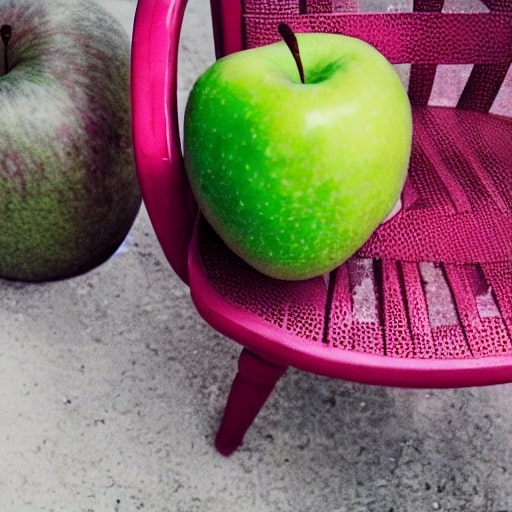}
        & 
    % SD
    \includegraphics[width=0.14\linewidth, height=0.08\textheight]{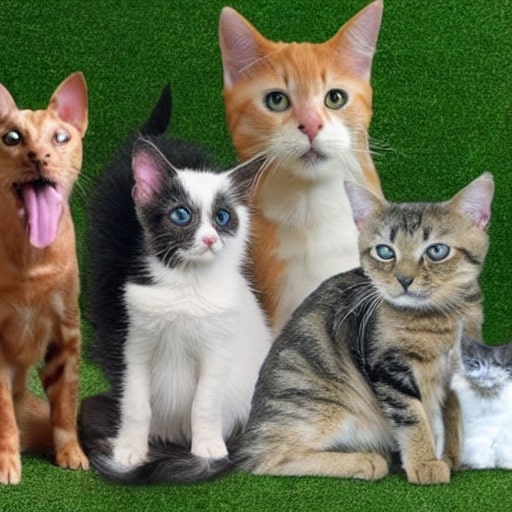}
	      &
    % AE
	\includegraphics[width=0.14\linewidth, height=0.08\textheight]{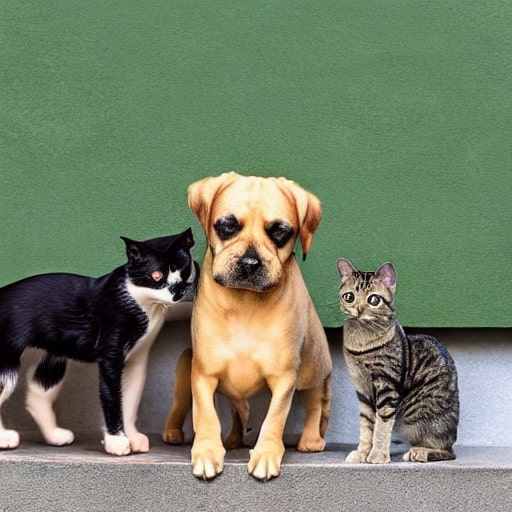}
	    & 
    % TV
	\includegraphics[width=0.14\linewidth, height=0.08\textheight]{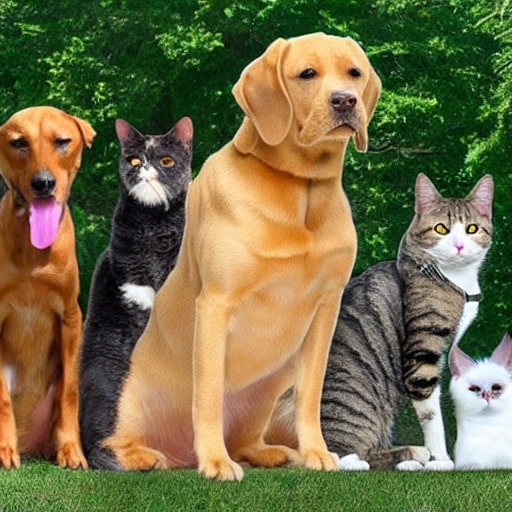}
    \vspace{-0.3em}
    \tabularnewline
    \footnotesize Stable Diffusion & 
    \footnotesize Attend\&Excite & 
    \footnotesize \textbf{{\ours}} &
    \footnotesize Stable Diffusion & 
    \footnotesize Attend\&Excite & 
    \footnotesize \textbf{{\ours}}
    \tabularnewline
	\end{tabular}
\hfill{}
\par\end{centering}
\caption{Limitations: challenging rare combinations (left) and instance miscounting (right).
} 
\label{fig:limitation}
\end{figure*}
Despite improved semantic guidance, it is yet difficult to generate extremely rare or implausible cases, e.g.,~unusual color binding \myquote{a gray apple}. %We observe that 
Our method may generate such objects together with the common one, e.g.,~generating a green apple and a gray apple in the same image, see \cref{fig:limitation}. As we use the pretrained model without fine-tuning, some data bias is inevitably inherited. 
Another issue is miscounting: more instances may be generated than it should. %, while the other methods tend to miss objects. 
We attribute the miscounting to the imprecise language understanding limited by the CLIP text encoder\newcite{radford2021clip,paiss2023teaching}. This effect is also observed in other large-scale T2I models, e.g., Parti\newcite{yu2022parti}, making it an interesting case for future research.

\section{Conclusion}\label{sec:conclusion}
In this work, we propose a novel inference-time optimization objective {\ours} for semantic nursing of pretrained T2I diffusion models.
Targeting at mitigating semantic issues in T2I synthesis, our approach demonstrates its effectiveness in generating multiple instances with correct attribute binding given complex textual descriptions.
We believe that our regularization technique can provide insights in the generation process and support further development in producing images semantically faithful to the textual input.

%%%%%%%%% REFERENCES
\bibliographystyle{iclr2023_conference}
\bibliography{reference}

%%%%%%%%% Appendix %%%%%%%%
\clearpage

\appendix
% reset the counter
\renewcommand{\thetable}{S.\arabic{table}}  
\renewcommand{\thefigure}{S.\arabic{figure}} 
\renewcommand{\theequation}{S.\arabic{equation}}
\renewcommand{\thesection}{S.\arabic{section}}
\refstepcounter{figure} 
\refstepcounter{table} 
\refstepcounter{equation}
\setcounter{figure}{0}  
\setcounter{table}{0}  
\setcounter{equation}{0}

\newcommand{\bmvtitlesize}{\fontsize{17}{17pt}\selectfont}
\def\bmv@strut{\rule{0pt}{1ex}}

{\LARGE\sc Supplementary Material \par}

% Overview
This supplementary material to the main paper is structured as follows:
\begin{itemize}
    \item In \cref{sec:appendix-visual}, more visual comparison is provided.
    \item In \cref{sec:appendix-numbers}, we provide additional quantitative evaluation using more metrics and with other methods.
    \item In \cref{sec:appendix-ablation}, we ablate on the binding loss $L_{bind}$.
    \item In \cref{sec:appendix-eval}, we present the algorithm overview, computation complexity and more details on the TIFA evaluation.
   % \item 
\end{itemize}
More attention visualization can be found in our \href{https://sites.google.com/view/divide-and-bind}{project page}.

\begin{figure*}[htb!]
\begin{centering}
\setlength{\tabcolsep}{0.0em}
\renewcommand{\arraystretch}{0}
\par\end{centering}
\begin{centering}
%\vspace{-2em}
\hfill{}%
	\begin{tabular}{
 @{\hspace{-0.25em}}c
 @{\hspace{0.1em}}c@{\hspace{0.2em}}c
 @{\hspace{0.6em}}c@{\hspace{0.2em}}c
 @{\hspace{0.6em}}c@{\hspace{0.2em}}c
 }
	\centering
    &
    \multicolumn{2}{c}{\begin{tabular}{c}\small 
    \myquote{The \textblue{flash} and  \\
    \small the \textblue{superman}\\
    \small on the snowy street
    }\end{tabular}} 
    & 
    \multicolumn{2}{c}{\begin{tabular}{c} \small 
    \myquote{The black \textblue{widow}  \\
    \small and the \textblue{spiderman}  \\
    \small on the beach
    }\end{tabular}} 
    &
    \multicolumn{2}{c}{\begin{tabular}{c} %\small 
    \small 
    \hspace{-0.2em}
    \myquote{
    The \textblue{flash} with green\\
    \small  suit  and the \textblue{batman} \\
    \small  with blue suit
    }\end{tabular}} 
	%\vspace{0.02cm} 
    %% Stable Diffsuion
    \tabularnewline
    \multirow{1}{*}{\rotatebox{90}{
        \hspace{4.5em}
        \begin{tabular}{c}
        \small Stable\\ \small Diffusion 
        \end{tabular}
        \hspace{-4.5em}
    }} 
        &
	\includegraphics[height= 0.09\textheight]{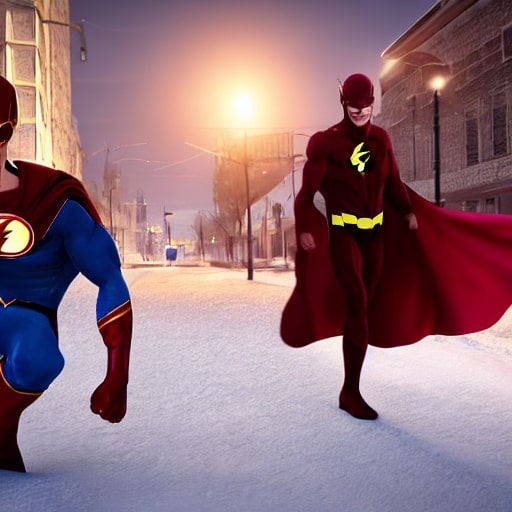}
	      &
	\includegraphics[height= 0.09\textheight]{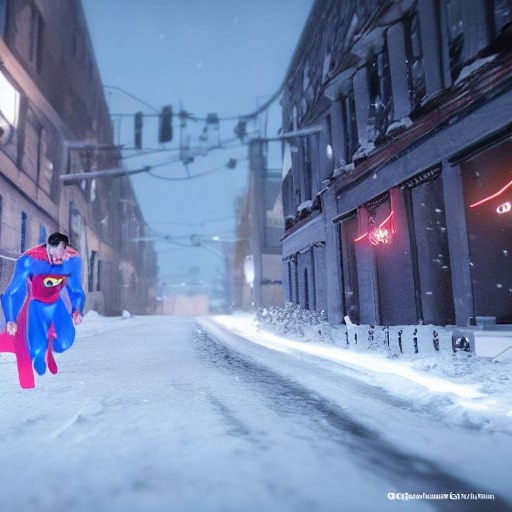}
	    & 
	\includegraphics[height= 0.09\textheight]{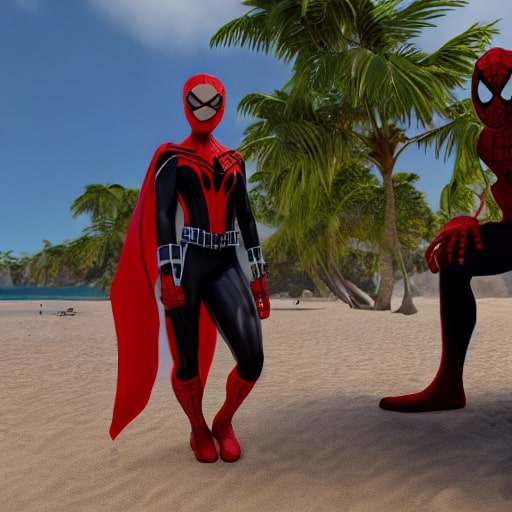}
        & 
    \includegraphics[height= 0.09\textheight]{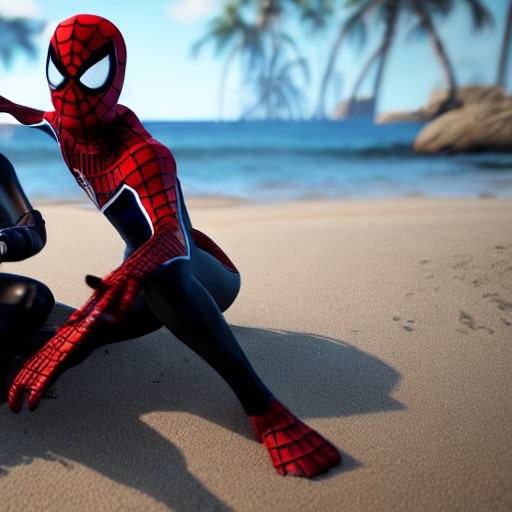}
	      & 
	\includegraphics[height= 0.09\textheight]{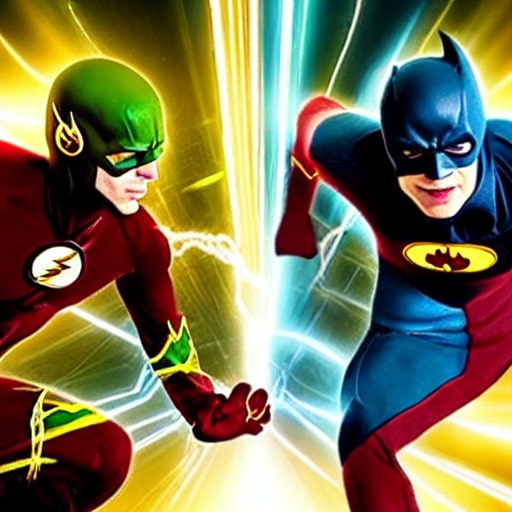}
	    & 
	\includegraphics[height= 0.09\textheight]{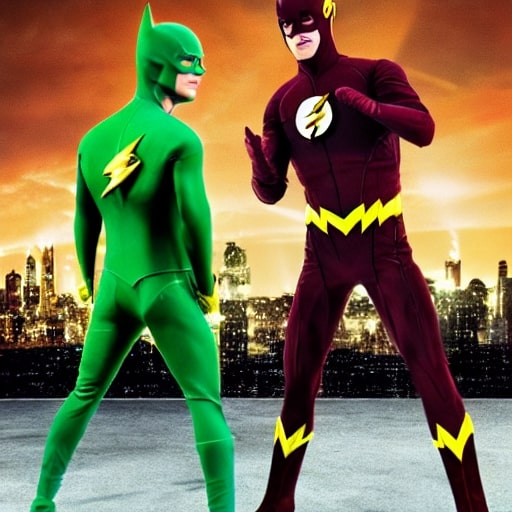}
	 %% Attend and Excite
  \tabularnewline
  \multirow{1}{*}{\rotatebox{90}{
        \hspace{4.5em}
        \begin{tabular}{c}
        \small Attend \& \\ \small Excite
        \end{tabular}
        \hspace{-4.5em}
    }} 
    &
    \includegraphics[height= 0.09\textheight]{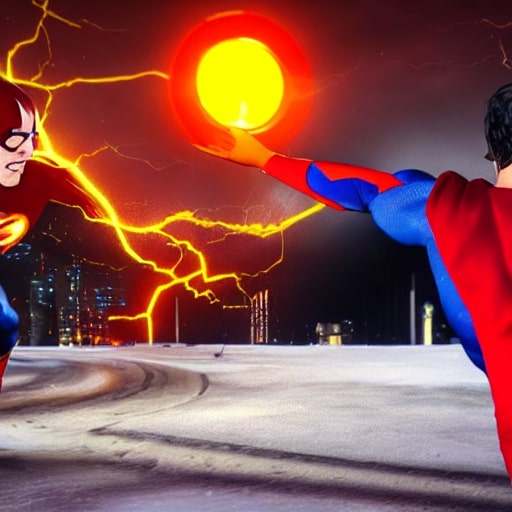}
	      &
	\includegraphics[height= 0.09\textheight]{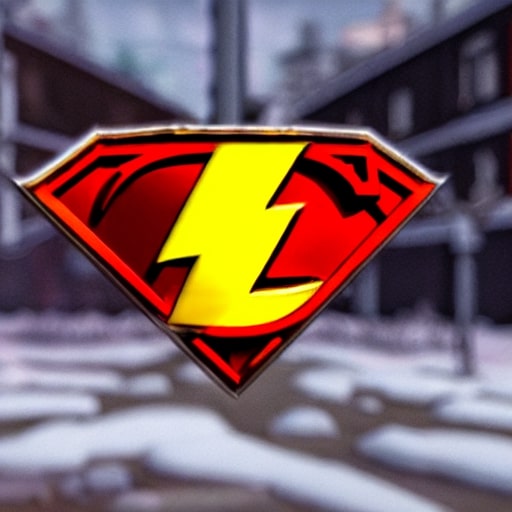}
	    & 
	\includegraphics[height= 0.09\textheight]{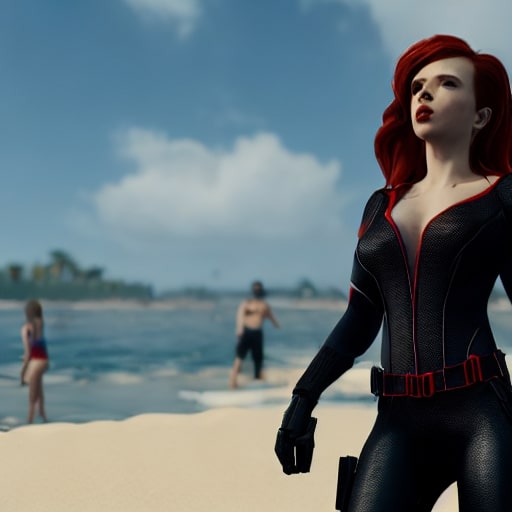}
        & 
    \includegraphics[height= 0.09\textheight]{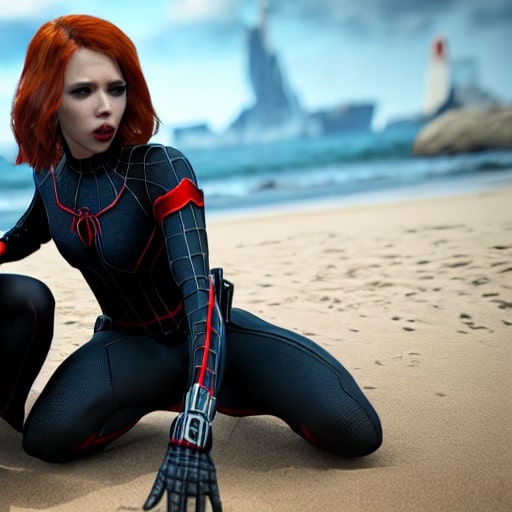}
	      & 
    \includegraphics[height= 0.09\textheight]{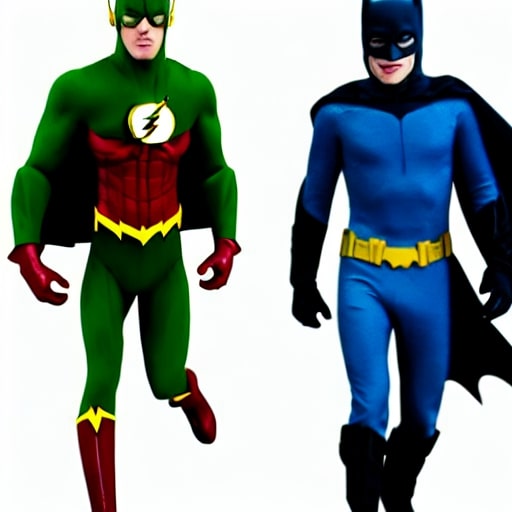}
	    & 
	\includegraphics[height= 0.09\textheight]{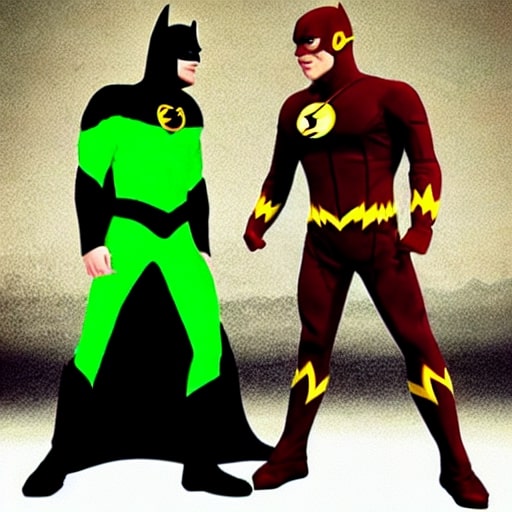}
	 \tabularnewline
  %% Ours
    \multirow{1}{*}{\rotatebox{90}{
        \hspace{4.5em}
        \begin{tabular}{c}
           \small \textbf{Divide \&} \\ \small \textbf{Bind} 
        \end{tabular}
        \hspace{-4.5em}
    }} 
    &
	\includegraphics[height= 0.09\textheight]{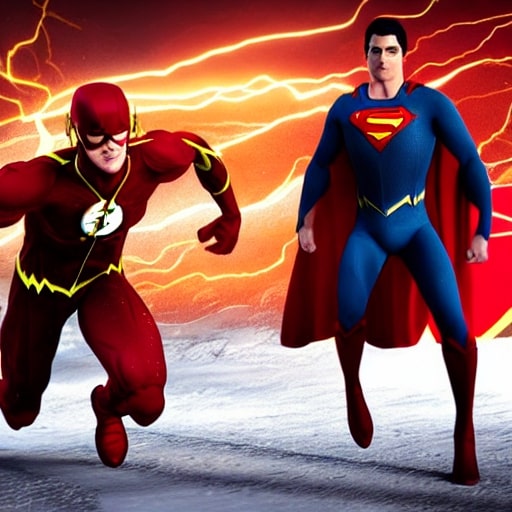}
	      &
	\includegraphics[height= 0.09\textheight]{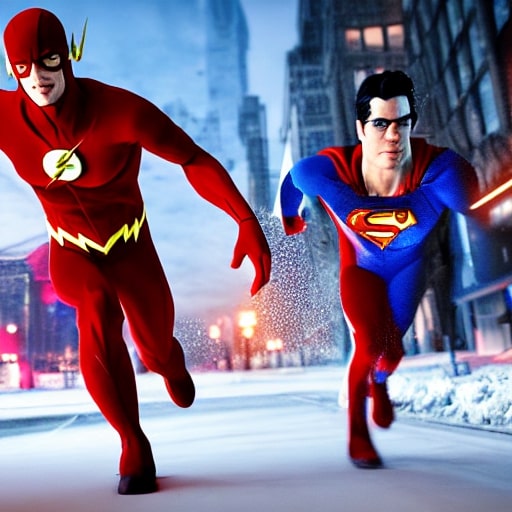}
	    & 
	\includegraphics[height= 0.09\textheight]{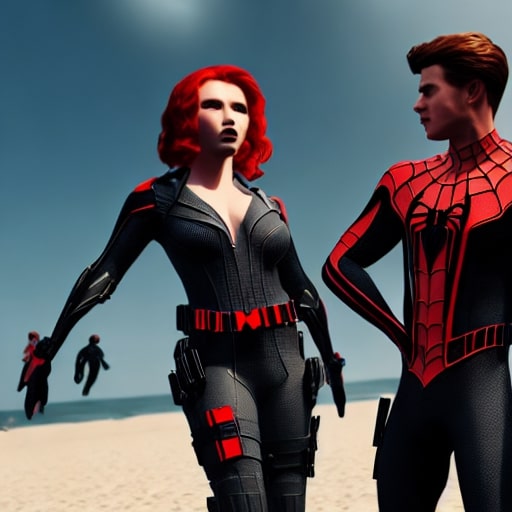}
        & 
    \includegraphics[height= 0.09\textheight]{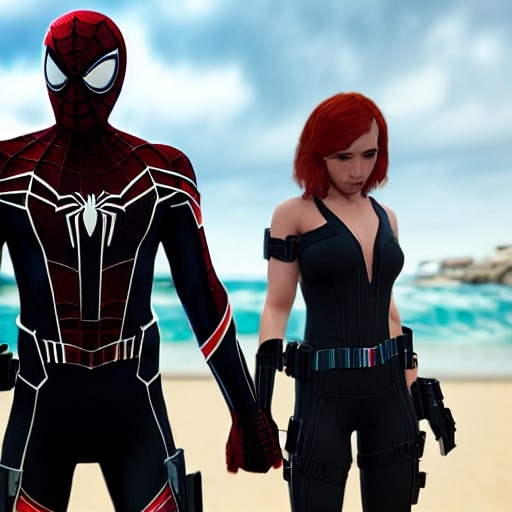}
	      & 
    \includegraphics[height= 0.09\textheight]{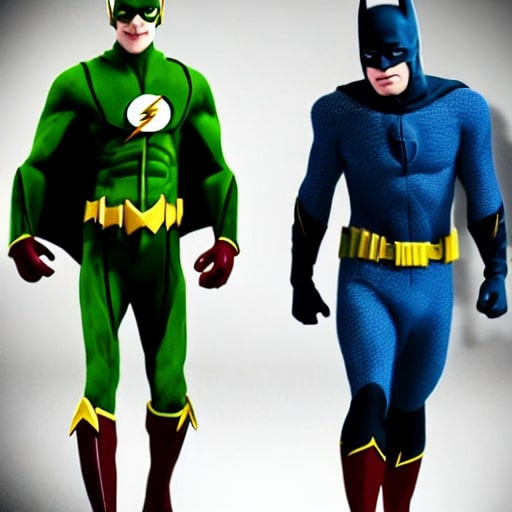}
	    & 
	\includegraphics[height= 0.09\textheight]{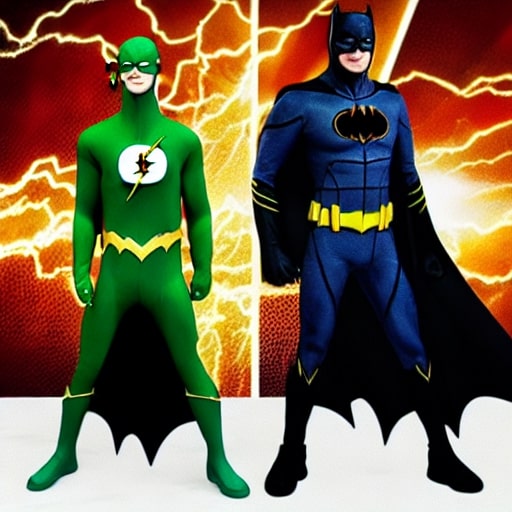}
    %\vspace{0.5em}
    \tabularnewline
	\end{tabular}
\hfill{}
\par\end{centering}
\caption{Qualitative comparison using novel prompts with the same random seeds. Tokens used for optimization are highlighted in \textblue{blue}. Compared to others, {\ours} can better comply with the input prompt while maintaining a high level of realism. 
} 
\label{fig:app_visual_compare_add}
\vspace{-0.5em}
\end{figure*}
%between Stable Diffusion, Attend and TV
%%%%%%%%%%%%%%%%%%%%%%%%%%%%%%%%%%%%%%%%%%%%
\section{Additional Qualitative Results}\label{sec:appendix-visual}

\begin{figure*}[h!]
\begin{centering}
\setlength{\tabcolsep}{0.0em}
\renewcommand{\arraystretch}{0}
\par\end{centering}
\begin{centering}
%\vspace{-2em}
\hfill{}%
	\begin{tabular}{
 @{\hspace{-0.25em}}c
 @{\hspace{0.1em}}c@{\hspace{0.2em}}c
 @{\hspace{0.6em}}c@{\hspace{0.2em}}c
 @{\hspace{0.6em}}c@{\hspace{0.2em}}c
 }
	\centering
    &
    \multicolumn{2}{c}{\begin{tabular}{c}\small 
    \myquote{A \textblue{dog} and a \textblue{turtle}
    }\end{tabular}} 
    & 
    \multicolumn{2}{c}{\begin{tabular}{c} \small 
    \myquote{A \textblue{dog}  and a \textblue{turtle} \\
    \small in the library
    }\end{tabular}} 
    &
    \multicolumn{2}{c}{\begin{tabular}{c} %\small 
    % \myquote{A white \textblue{firehydrant} \\ 
    % \small  on the sidewalk \\ 
    % \small by a black \textblue{car}
    \small 
    \hspace{-0.2em}
    \myquote{A \textblue{dog}  and a \textblue{turtle} on \\ 
    \small the street, snowy scene
    }\end{tabular}} 
	%\vspace{0.02cm} 
    %% Stable Diffsuion
    \tabularnewline
    \multirow{1}{*}{\rotatebox{90}{
        \hspace{4.5em}
        \begin{tabular}{c}
        \small Stable\\ \small Diffusion 
        \end{tabular}
        \hspace{-4.5em}
    }} 
        &
	\includegraphics[height= 0.09\textheight]{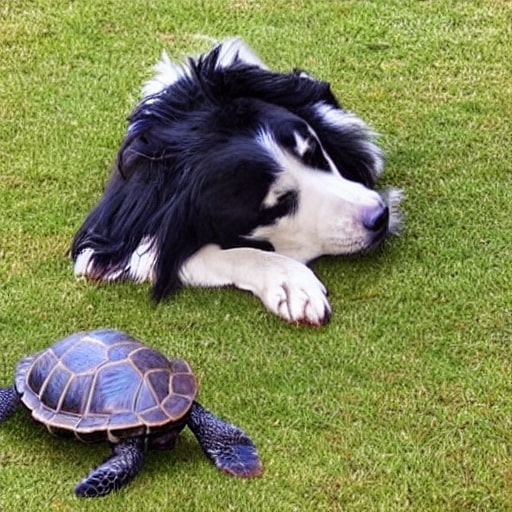}
	      &
	\includegraphics[height= 0.09\textheight]{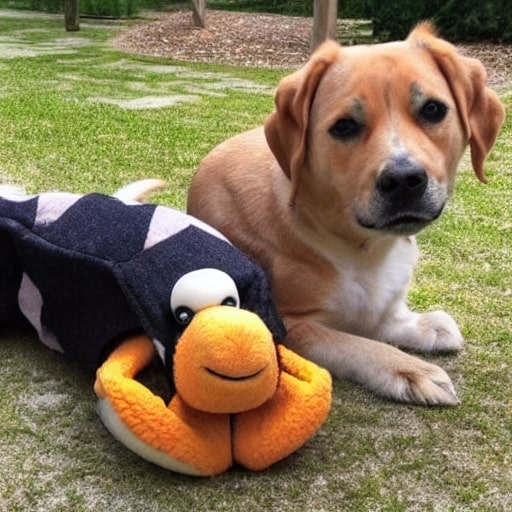}
	    & 
	\includegraphics[height= 0.09\textheight]{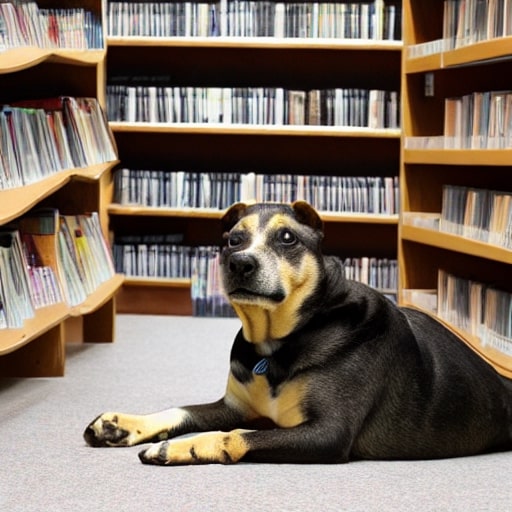}
        & 
    \includegraphics[height= 0.09\textheight]{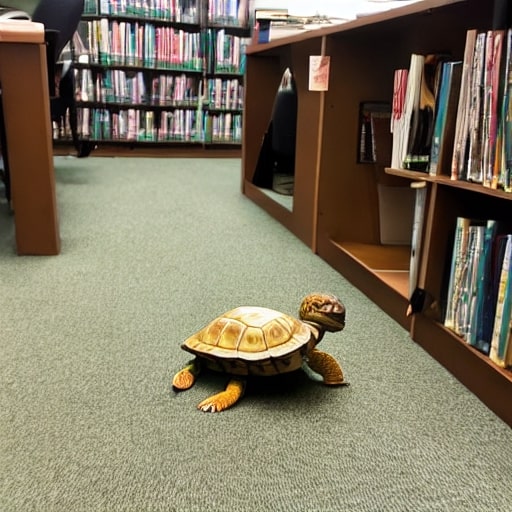}
	      & 
	\includegraphics[height= 0.09\textheight]{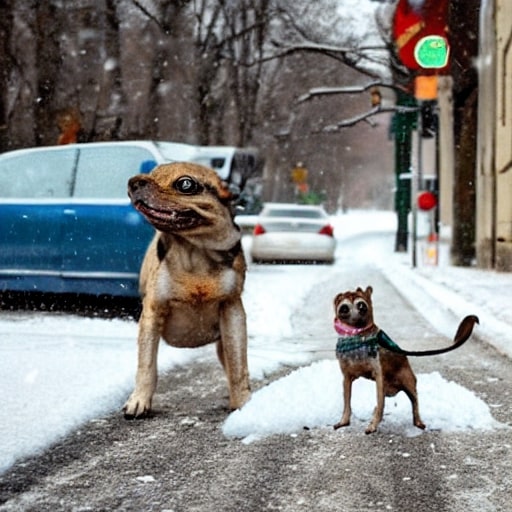}
	    & 
	\includegraphics[height= 0.09\textheight]{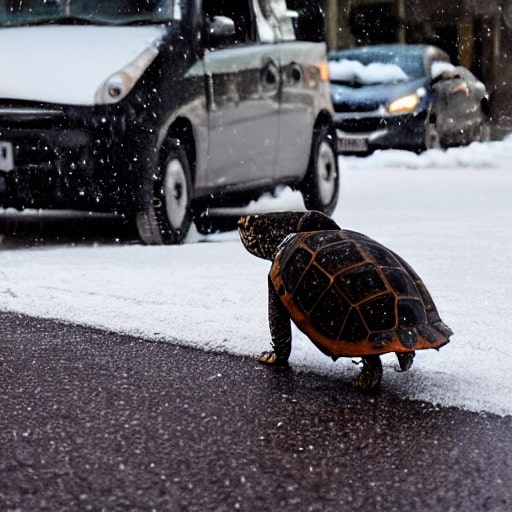}
	 %% Attend and Excite
  \tabularnewline
  \multirow{1}{*}{\rotatebox{90}{
        \hspace{4.5em}
        \begin{tabular}{c}
        \small Attend \& \\ \small Excite
        \end{tabular}
        \hspace{-4.5em}
    }} 
    &
    \includegraphics[height= 0.09\textheight]{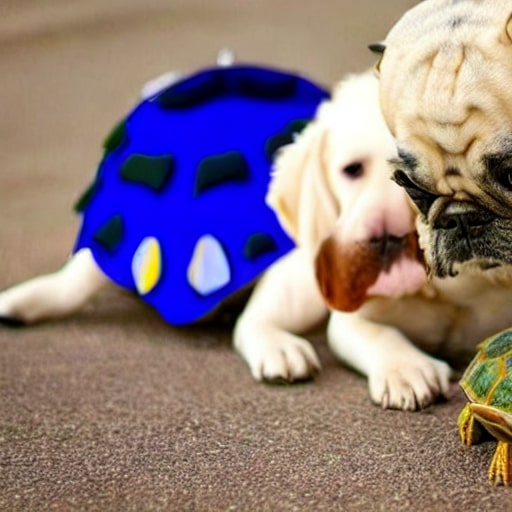}
	      &
	\includegraphics[height= 0.09\textheight]{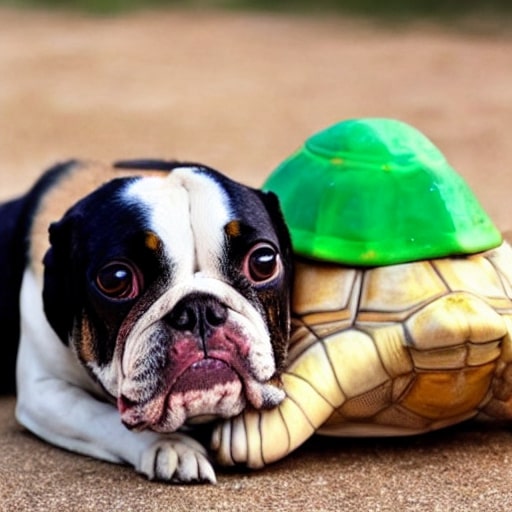}
	    & 
	\includegraphics[height= 0.09\textheight]{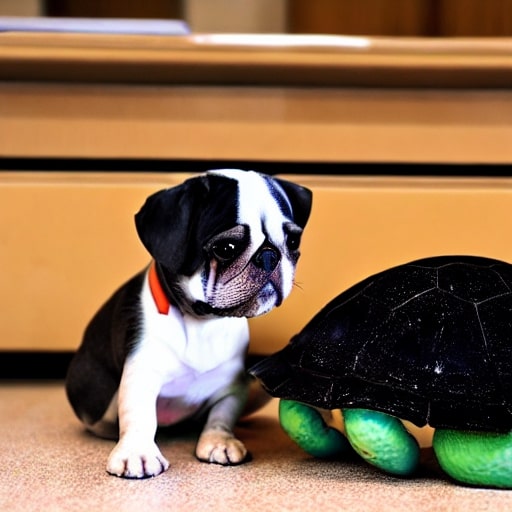}
        & 
    \includegraphics[height= 0.09\textheight]{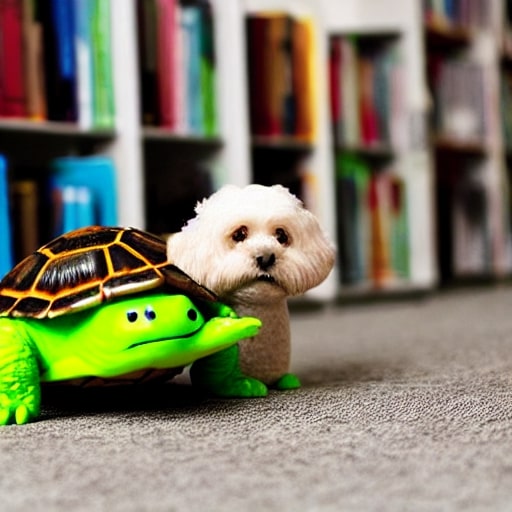}
	      & 
    \includegraphics[height= 0.09\textheight]{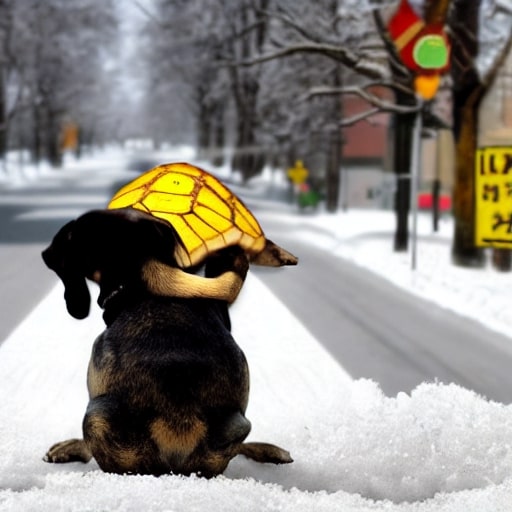}
	    & 
	\includegraphics[height= 0.09\textheight]{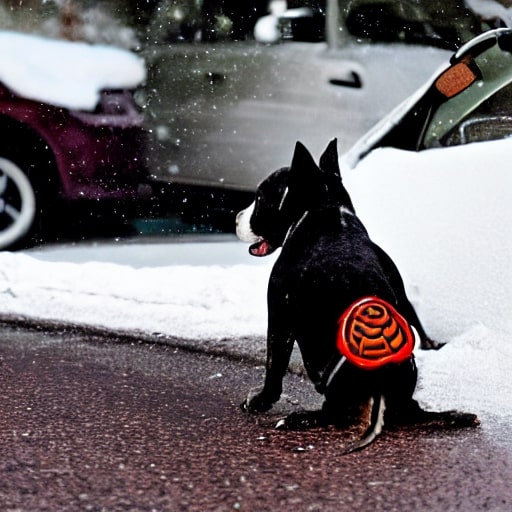}
	 \tabularnewline
  %% Ours
    \multirow{1}{*}{\rotatebox{90}{
        \hspace{4.5em}
        \begin{tabular}{c}
           \small \textbf{Divide \&} \\ \small \textbf{Bind} 
        \end{tabular}
        \hspace{-4.5em}
    }} 
    &
	\includegraphics[height= 0.09\textheight]{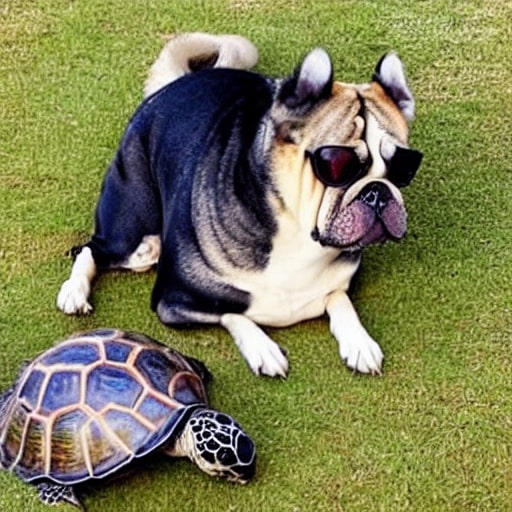}
	      &
	\includegraphics[height= 0.09\textheight]{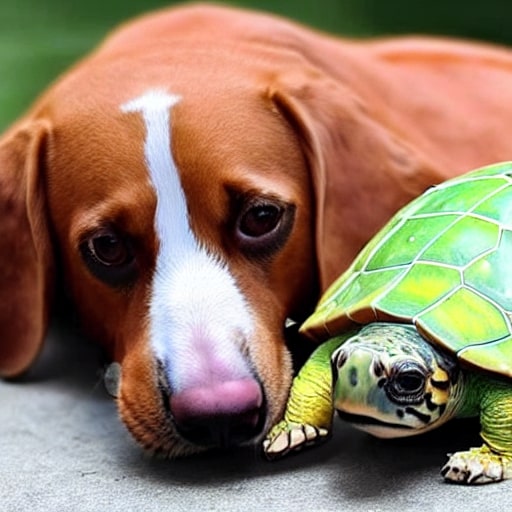}
	    & 
	\includegraphics[height= 0.09\textheight]{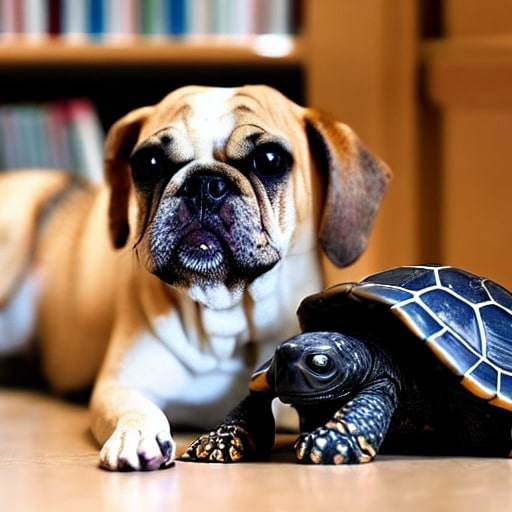}
        & 
    \includegraphics[height= 0.09\textheight]{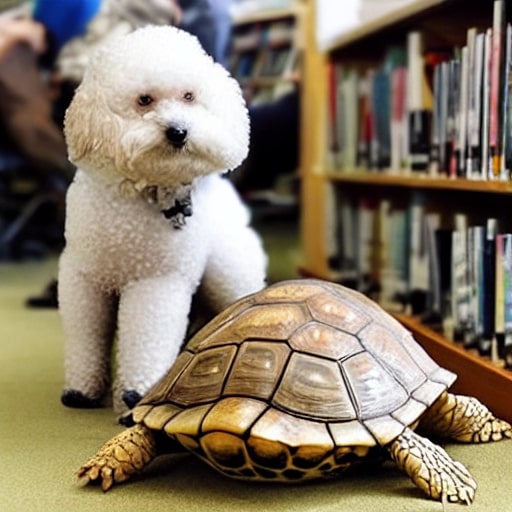}
	      & 
    \includegraphics[height= 0.09\textheight]{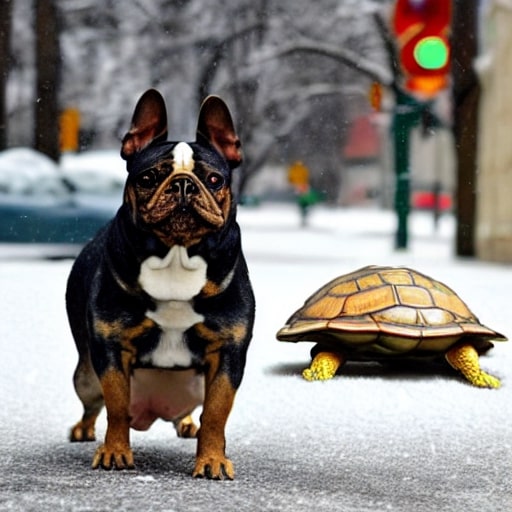}
	    & 
	\includegraphics[height= 0.09\textheight]{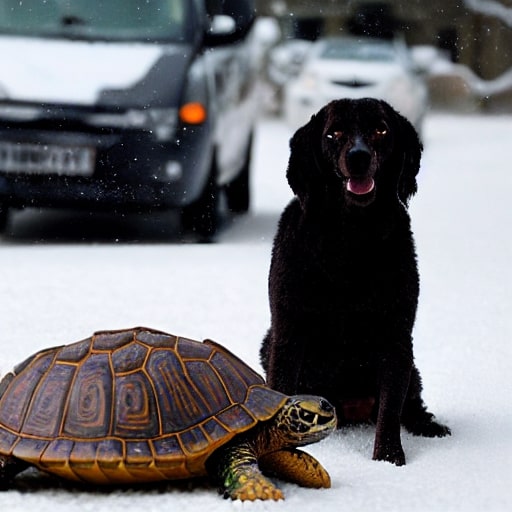}
    \vspace{2em}
    \tabularnewline
        &
    \multicolumn{2}{c}{\begin{tabular}{c}\small 
    \myquote{A red sports \textblue{car} \\ 
    \small is parked \\ 
    \small beside a black \textblue{horse} 
    }\end{tabular}} 
    & 
    \multicolumn{2}{c}{\begin{tabular}{c} \small 
    \myquote{A blue \textblue{dog}  \\ 
    \small on a red \textblue{coach}
    }\end{tabular}} 
    &
    \multicolumn{2}{c}{\begin{tabular}{c} %\small 
    % \myquote{A white \textblue{firehydrant} \\ 
    % \small  on the sidewalk \\ 
    % \small by a black \textblue{car}
    \small 
    \myquote{A brown \textblue{dog} \\ 
    \small  sitting in the yard \\ 
    \small with a white \textblue{cat}
    }\end{tabular}} 
	%\vspace{0.02cm} 
    %% Stable Diffsuion
    \tabularnewline
    \multirow{1}{*}{\rotatebox{90}{
        \hspace{4.5em}
        \begin{tabular}{c}
        \small Stable\\ \small Diffusion 
        \end{tabular}
        \hspace{-4.5em}
    }} 
        &
	\includegraphics[height= 0.09\textheight]{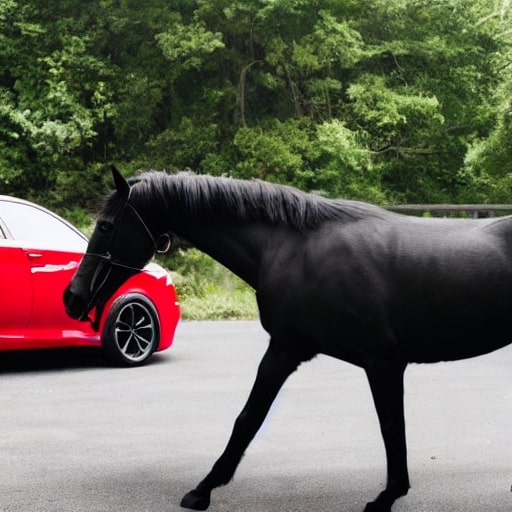}
	      &
	\includegraphics[height= 0.09\textheight]{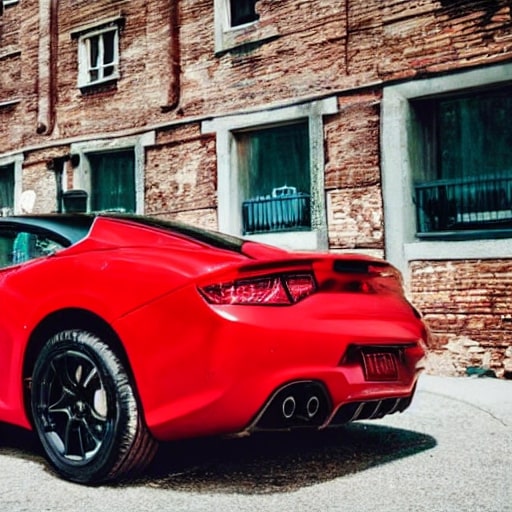}
	    & 
	\includegraphics[height= 0.09\textheight]{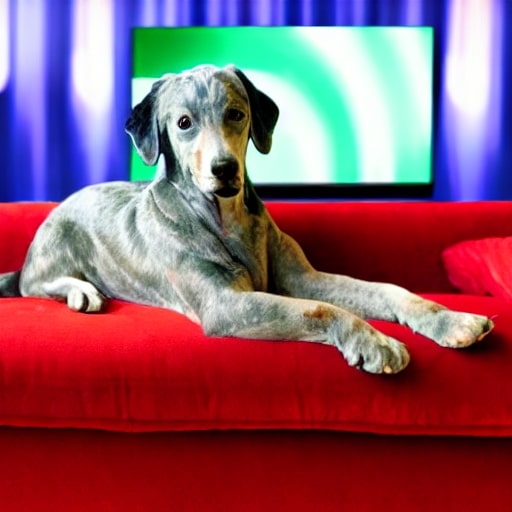}
        & 
    \includegraphics[height= 0.09\textheight]{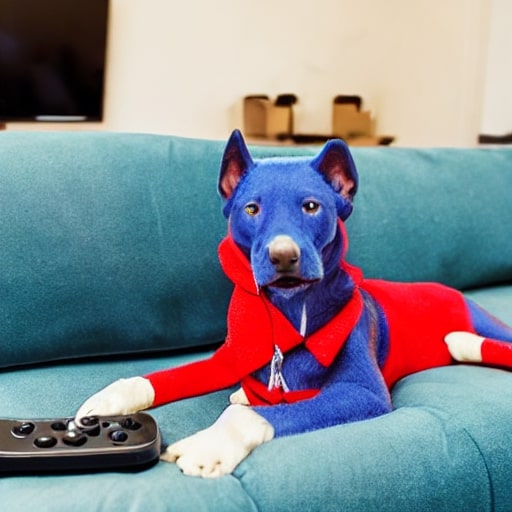}
	      & 
	%\includegraphics[height= 0.09\textheight]{appendix/fig/white_fire_black_car/18_5362_sd.jpg}
	%    & 
	%\includegraphics[height= 0.09\textheight]{appendix/fig/white_fire_black_car/48_967_sd.jpg}
	\includegraphics[height= 0.09\textheight]{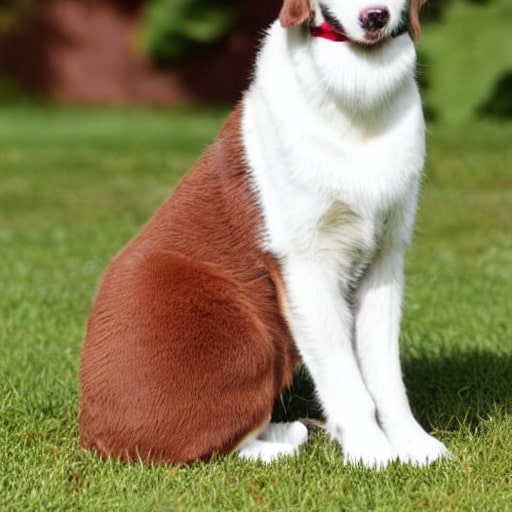}
	    & 
	\includegraphics[height= 0.09\textheight]{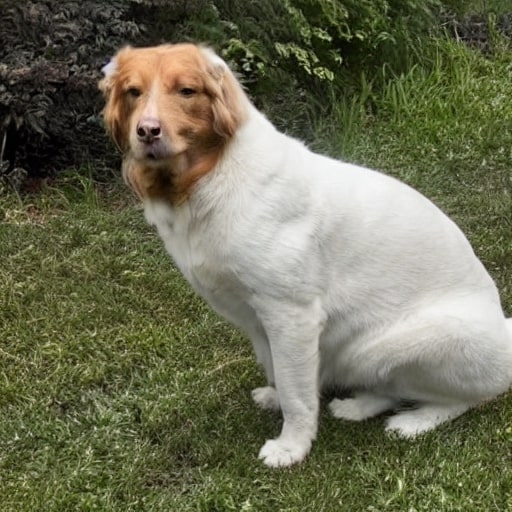}
	 %% Attend and Excite
  \tabularnewline
  \multirow{1}{*}{\rotatebox{90}{
        \hspace{4.5em}
        \begin{tabular}{c}
        \small Attend \& \\ \small Excite
        \end{tabular}
        \hspace{-4.5em}
    }} 
    &
    \includegraphics[height= 0.09\textheight]{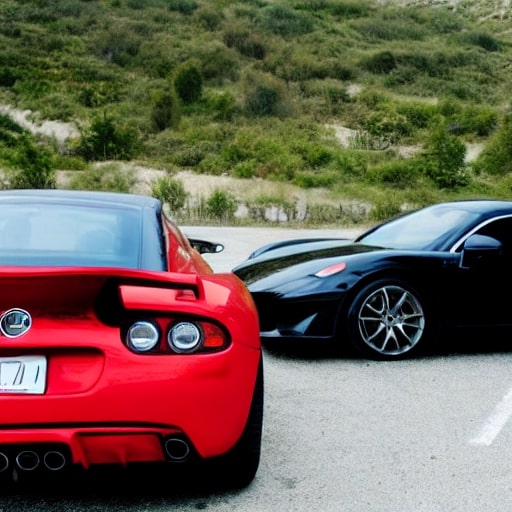}
	      &
	\includegraphics[height= 0.09\textheight]{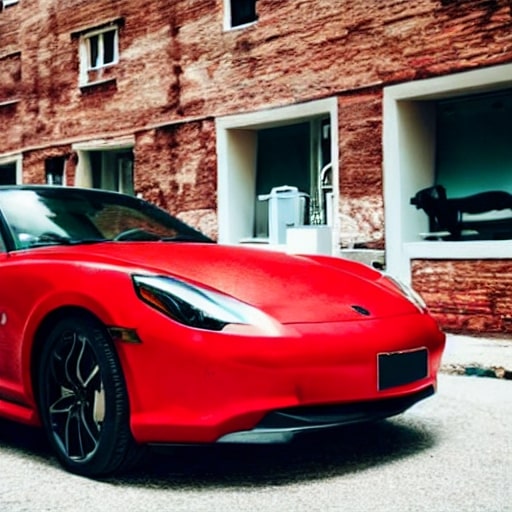}
	    & 
	\includegraphics[  height= 0.09\textheight]{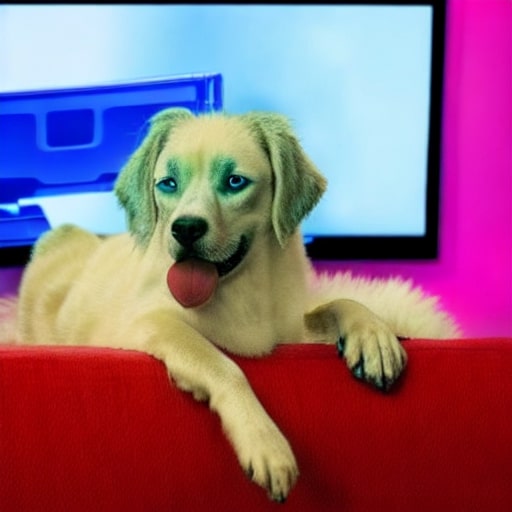}
        & 
    \includegraphics[height= 0.09\textheight]{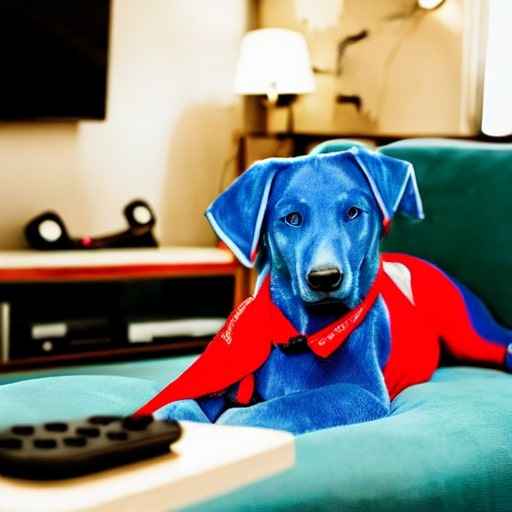}
	      & 
% 	\includegraphics[height= 0.09\textheight]{appendix/fig/white_fire_black_car/18_5362_excite.jpg}
% 	    & 
% 	\includegraphics[height= 0.09\textheight]{appendix/fig/white_fire_black_car/48_967_excite.jpg}
    \includegraphics[height= 0.09\textheight]{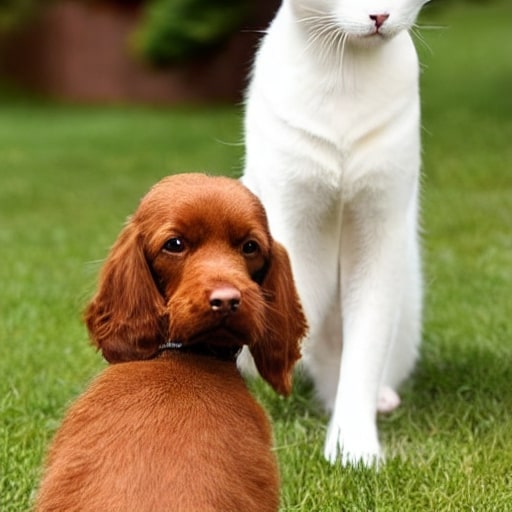}
	    & 
	\includegraphics[height= 0.09\textheight]{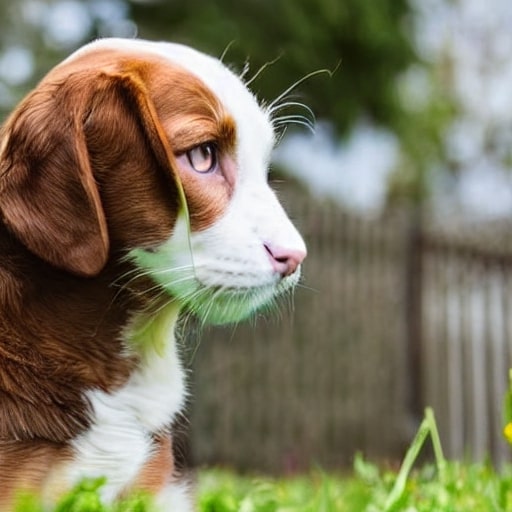}
	 \tabularnewline
  %% Ours
    \multirow{1}{*}{\rotatebox{90}{
        \hspace{4.5em}
        \begin{tabular}{c}
           \small \textbf{Divide \&} \\ \small \textbf{Bind} 
        \end{tabular}
        \hspace{-4.5em}
    }} 
    &
	\includegraphics[height= 0.09\textheight]{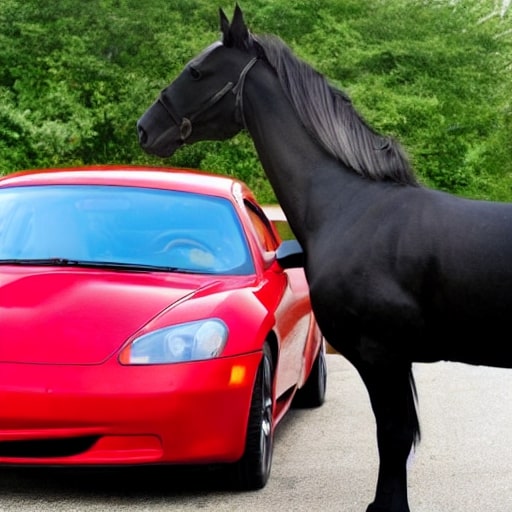}
	      &
	\includegraphics[height= 0.09\textheight]{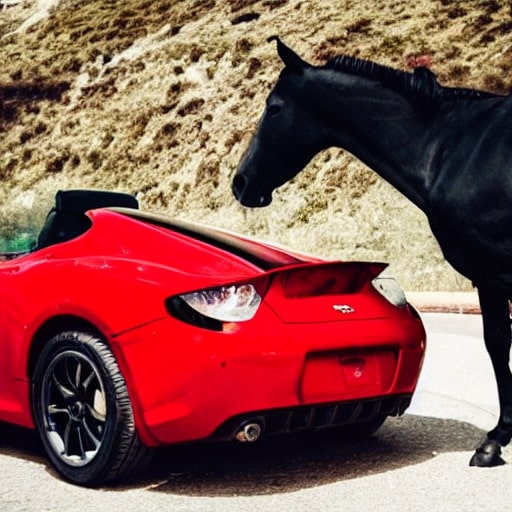}
	    & 
	\includegraphics[height= 0.09\textheight]{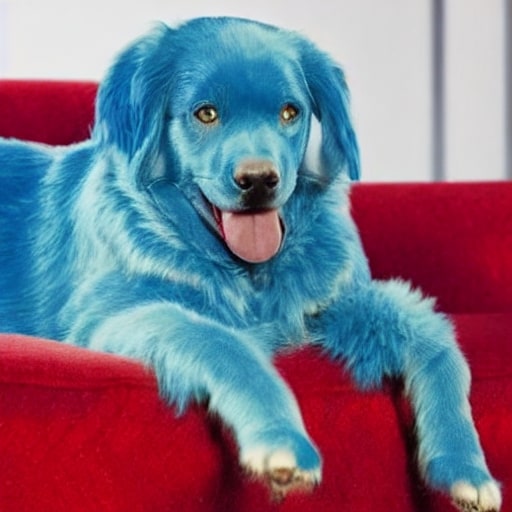}
        & 
    \includegraphics[height= 0.09\textheight]{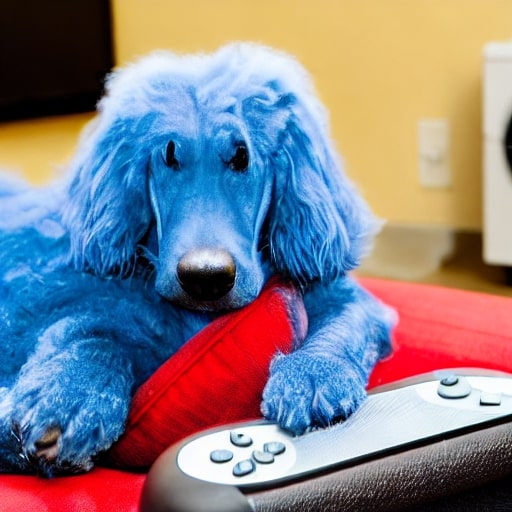}
	      & 
% 	\includegraphics[height= 0.09\textheight]{appendix/fig/white_fire_black_car/18_5362_tv-bind-v4-maxIt25.jpg}
% 	    & 
% 	\includegraphics[height= 0.09\textheight]{appendix/fig/white_fire_black_car/48_967_tv-bind-v4-maxIt25.jpg}
    \includegraphics[height= 0.09\textheight]{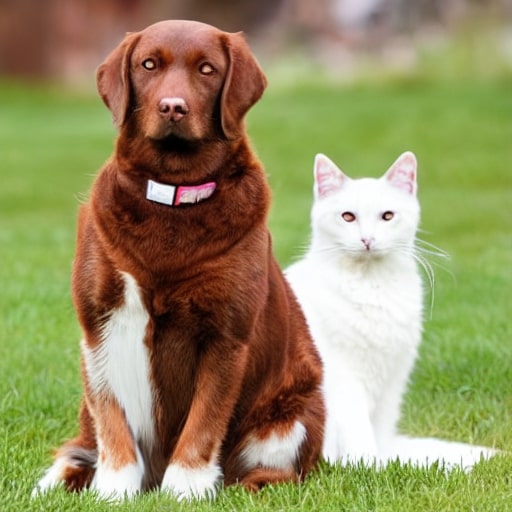}
	    & 
	\includegraphics[height= 0.09\textheight]{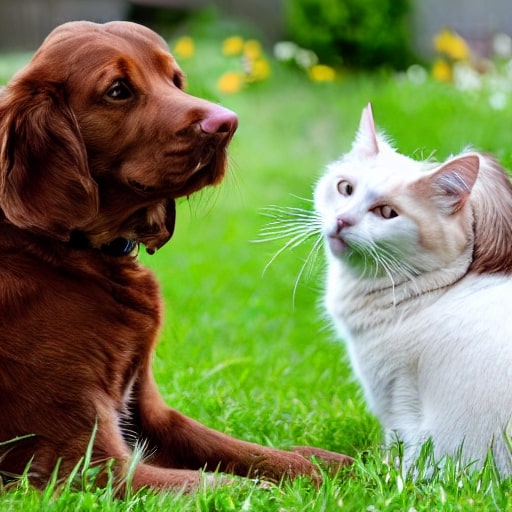}
    %\vspace{0.5em}
    \tabularnewline
	\end{tabular}
\hfill{}
\par\end{centering}
\caption{Qualitative comparison in different settings with the same random seeds. Tokens used for optimization are highlighted in \textblue{blue}. Compared to others, {\ours} shows superior alignment with the input prompt while maintaining a high level of realism.
} 
\label{fig:app_visual_compare_animal}
%\vspace{-0.2em}
\end{figure*}

We provide more visual comparison using additional novel prompts in \cref{fig:app_visual_compare_add}  and across different benchmarks using the same random seed in \cref{fig:app_visual_compare_animal}. As can be seen, {\ours} can handle various complex prompts well and outperform the other methods in different scenarios.

% More visual 
%between Stable Diffusion, Attend and TV
%\input{appendix/fig/visual_tifa_attr} % merged

%%%%%%%%%%%%%%%%%%%%%%%%%%%%%%%%%%%%%%%%%%%%
\section{Additional Quantitative Evaluation} \label{sec:appendix-numbers}

In \cref{tab:app_clip_sim}, we compare our {\ours} with Stable Diffusion and Attend \& Excite using Full Prompt similarity and Minimum Object Similarity used in \newcite{chefer2023attendandexcite}. Full Prompt Similarity represents the average CLIP cosine similarity between the full text prompt and the generated images. And the Minimum Object Similarity is the minimum value of the object CLIP similarity among all objects mentioned in the prompt. For instance, for the prompt \myquote{a cat and a dog}, we compute the similarity between the image and the sub-phrase \myquote{a dog} and  \myquote{a cat} and take the smaller value as the final result. 
The difference among methods using CLIP similarities are minor, due to the fact that CLIP similarity may not be accurate to evaluate the faithfulness of Text-to-Image synthesis\newcite{hu2023tifa,Lu2023LLMScoreUT}. Therefore, we employed more recent evaluate metrics, TIFA score\newcite{hu2023tifa} and Text-Text similarity, 
for more reliable evaluation, as reported in Fig. \textcolor{red}{6} and Table \textcolor{red}{2} in the main paper.

\begin{table}[t]
\setlength{\tabcolsep}{0.5em}
\renewcommand{\arraystretch}{1.3}
\begin{center}
{\footnotesize  %\small\footnotesize 
 %\vspace{-1.4em}
%\begin{minipage}{.4\linewidth}
% text-text and tifa
    \begin{tabular}{@{\extracolsep{2pt}}l  c c @{\hspace{1em}} c c @{\hspace{1em}} c  c @{}}
    %\toprule
    \multirow{2}{*}{Method}  &
    \multicolumn{2}{c}{{Animal-Animal}} &
    \multicolumn{2}{c}{{Animal-Scene}}  &  \multicolumn{2}{c}{{\tifaCOCO}}  \\  
    \cline{2-3} \cline{4-5} \cline{6-7}
    & Full Prompt & Min. Obj. & Full Prompt & Min. Obj. & Full Prompt & Min. Obj.\\
    \hline
    Stable Diffusion 
        & 0.312 & 0.220 & 0.348 & 0.206 & 0.324 & 0.229  \\
    Attend \& Excite  
        & 0.333 & 0.249 & 0.344 & 0.240 & 0.328 & 0.236  \\
    {\ours}
        & 0.331 & 0.246 & 0.345 & 0.236 & 0.329 & 0.236\\
    \end{tabular}
}
%\vspace{-1.0em}
\end{center} 
%\vspace{-0.4em}
\caption{
    Quantitative comparison using Full Prompt Similarity and Minimum Object Similarity. The differences between methods are minor, which may due to the suboptimality of the evaluation metric as pointed in\newcite{hu2023tifa}.
    %on complex COCO-captions and {Multi-Object} generation. {\ours} surpasses the other methods when it comes to handling complex prompts.
}
\label{tab:app_clip_sim}
%\vspace{-0.7em}
\end{table}

\begin{table}[t]
\setlength{\tabcolsep}{0.5em}
\renewcommand{\arraystretch}{1.3}
\begin{center}
{\footnotesize  %\small\footnotesize 
 %\vspace{-1.4em}
%\begin{minipage}{.4\linewidth}
% text-text and tifa
    \begin{tabular}{@{\extracolsep{2pt}}l  c c @{}}
    %\toprule
    Method  &
    {Animal-Animal} &
    {Color-Object}   \\  
    \hline
    Stable Diffusion\newcite{rombach2022SD} & 0.77 & 0.77  \\
    Composable Diffusion\newcite{liu2022compositional} & 0.69 & 0.76  \\
    Structure Diffusion\newcite{feng2023structureDiffusion} & 0.76 & 0.76  \\
    Attend \& Excite\newcite{chefer2023attendandexcite} & 0.80 & 0.81 \\
    {\ours} & \textbf{0.81} & \textbf{0.82}
    \end{tabular}
}
%\vspace{-1.0em}
\end{center} 
%\vspace{-0.4em}
\caption{
    Comparison with other Text-to-Image methods in Text-Text similarity.
    %on complex COCO-captions and {Multi-Object} generation. 
    {\ours} surpasses the other methods on both evaluation sets. 
}
\label{tab:app_text}
%\vspace{-0.7em}
\end{table}

In \cref{tab:app_text}, we additionally compare with two more text-to-image methods, Composable Diffusion\newcite{liu2022compositional} and Structure Diffusion\newcite{feng2023structureDiffusion} using Text-Text similarity.  We outperform the other methods on both Animal-Animal and Color-Object benchmarks.

%%%%%%%%%%%%%%%%%%%%%%%%%%%%%%%%%%%%%%%%%%%%
\section{Ablation Study} \label{sec:appendix-ablation}
\begin{figure*}[htb!]
\begin{centering}
\setlength{\tabcolsep}{0.0em}
\renewcommand{\arraystretch}{0}
\par\end{centering}
\begin{centering}
%\vspace{-2em}
\hfill{}%
	\begin{tabular}{
 @{\hspace{-0.25em}}c
 @{\hspace{0.1em}}c@{\hspace{0.2em}}c
 @{\hspace{0.6em}}c@{\hspace{0.2em}}c
 @{\hspace{0.6em}}c@{\hspace{0.2em}}c
 }
	\centering
    &
    \multicolumn{2}{c}{\begin{tabular}{c}\small 
    \myquote{A purple \textblue{dog} and a \\ 
    \small  green \textblue{bench} on the  \\
    \small street, snowy scene
    }\end{tabular}} 
    & 
    \multicolumn{2}{c}{\begin{tabular}{c} \small 
    \myquote{A green \textblue{balloon} and a \\ 
    \small  pink \textblue{car} on the street, \\
    \small  nighttime scene
    }\end{tabular}} 
    &
    \multicolumn{2}{c}{\begin{tabular}{c} %\small 
    % \myquote{A white \textblue{firehydrant} \\ 
    % \small  on the sidewalk \\ 
    % \small by a black \textblue{car}
    \small 
    \hspace{-0.2em}
    \myquote{A yellow \textblue{glasses}  \\
    \small and a gray \textblue{bowl} 
    }\end{tabular}} 
	%\vspace{0.02cm} 
    %% Without JSD
    \tabularnewline
    \multirow{1}{*}{\rotatebox{90}{
        \hspace{4.5em}
        \begin{tabular}{c}
        \small  w/o\ $L_{bind}$
        \end{tabular}
        \hspace{-4.5em}
    }} 
        &
	\includegraphics[height= 0.09\textheight]{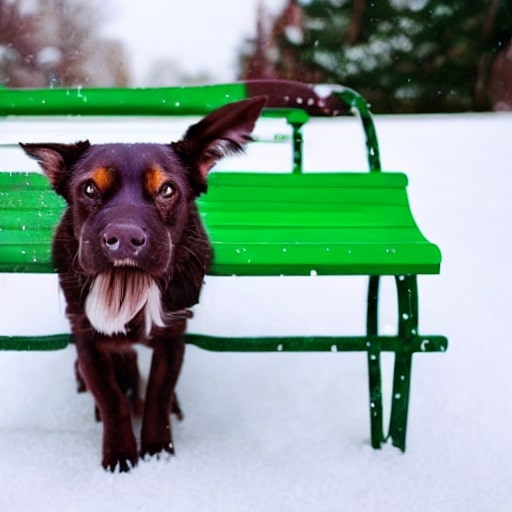}
	      &
	\includegraphics[height= 0.09\textheight]{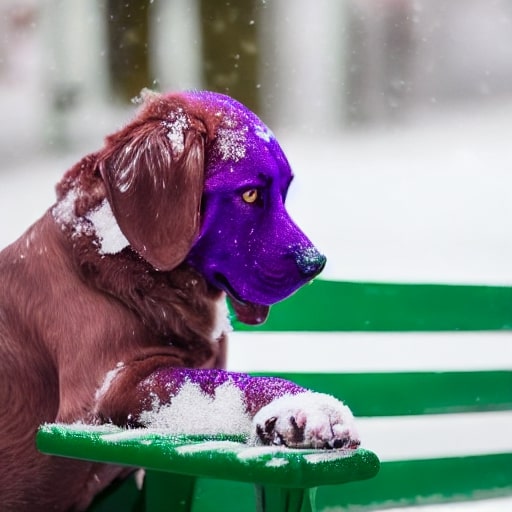}
	    & 
	\includegraphics[height= 0.09\textheight]{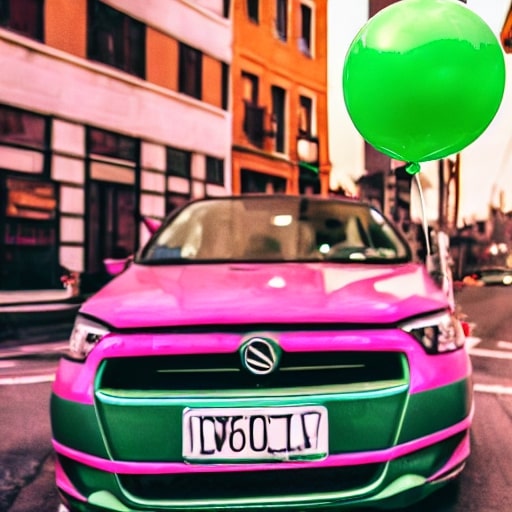}
        & 
    \includegraphics[height= 0.09\textheight]{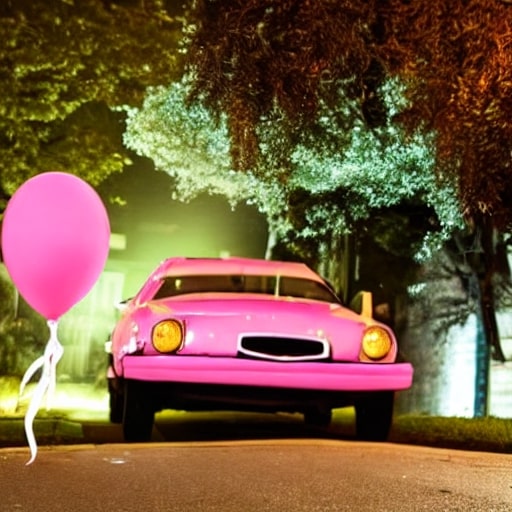}
	      & 
	\includegraphics[height= 0.09\textheight]{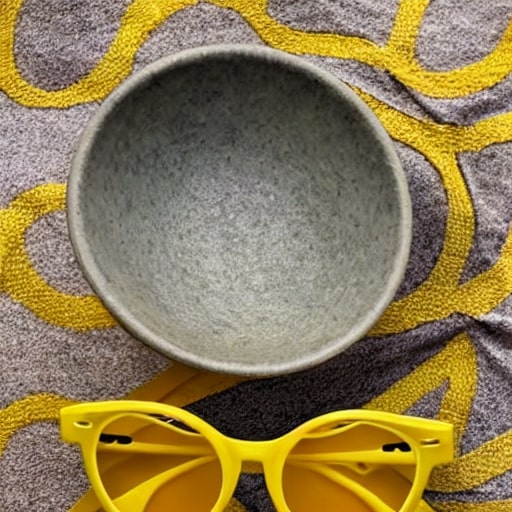}
	    & 
	\includegraphics[height= 0.09\textheight]{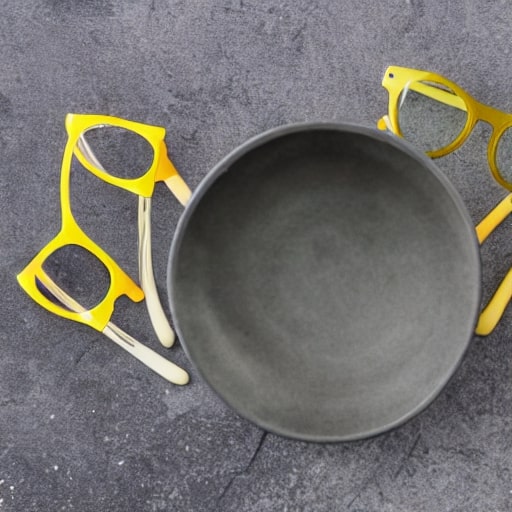}
    
    %% With JSD
  \tabularnewline
  \multirow{1}{*}{\rotatebox{90}{
        \hspace{4.5em}
        \begin{tabular}{c}
        \small w/- \ $L_{bind}$
        \end{tabular}
        \hspace{-4.5em}
    }} 
    &
    \includegraphics[height= 0.09\textheight]{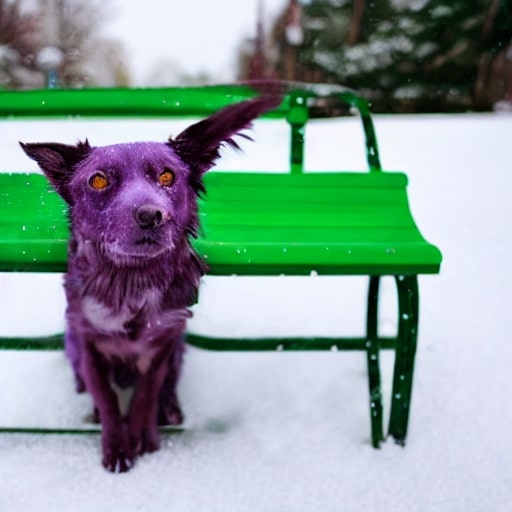}
	      &
	\includegraphics[height= 0.09\textheight]{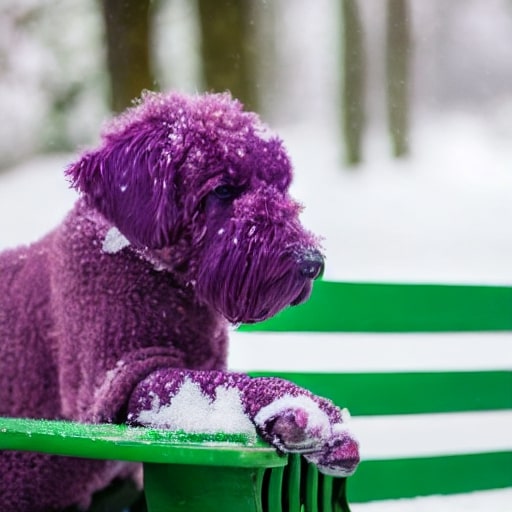}
	    & 
	\includegraphics[height= 0.09\textheight]{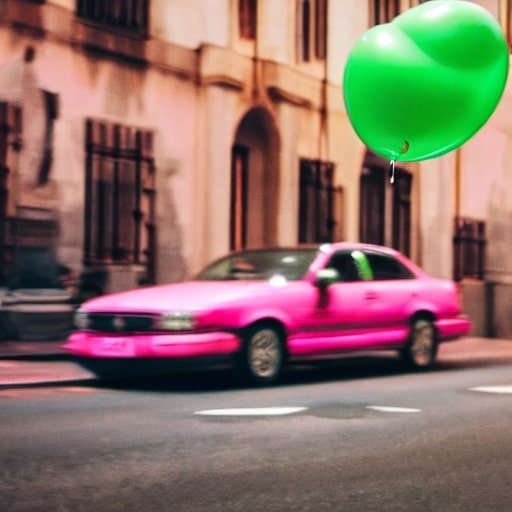}
        & 
    \includegraphics[height= 0.09\textheight]{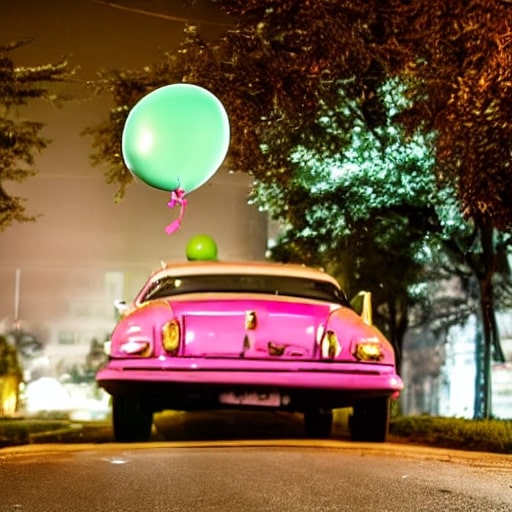}
	      & 
    \includegraphics[height= 0.09\textheight]{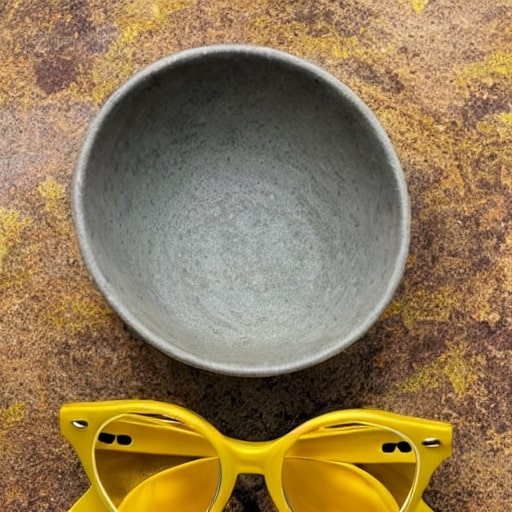}
	    & 
	\includegraphics[height= 0.09\textheight]{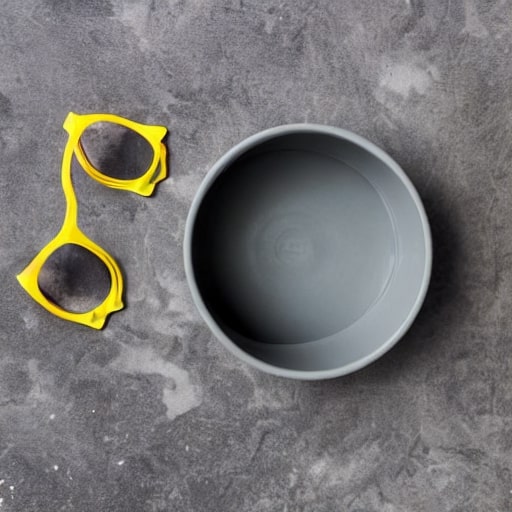}
	 \tabularnewline
	\end{tabular}
\hfill{}
\par\end{centering}
\caption{Qualitative ablation on the binding loss $L_{bind}$. With the binding loss, the attribute can be more accurately assigned to the corresponding object.
} 
\label{fig:app_visual_jsd}
%\vspace{-0.2em}
\end{figure*}
%between Stable Diffusion, Attend and TV
\begin{table}[t]
\setlength{\tabcolsep}{0.5em}
\renewcommand{\arraystretch}{1.3}
\begin{center}
{\footnotesize  %\small\footnotesize 
 %\vspace{-1.4em}
%\begin{minipage}{.4\linewidth}
% text-text and tifa
    \begin{tabular}{@{\extracolsep{4pt}}l  c c @{\hspace{1em}} c c @{\hspace{1em}} c  c @{}}
    %\toprule
    \multirow{2}{*}{Method}  &
    \multicolumn{2}{c}{{Color-Object}} &
    \multicolumn{2}{c}{{\colorScene}}  &  \multicolumn{2}{c}{{\tifaCOCO}}  \\  
    \cline{2-3} \cline{4-5} \cline{6-7}
    & Text-Text & TIFA & Text-Text & TIFA & Text-Text & TIFA \\
    \hline
    w/o $L_{bind}$ 
        & 0.815 & 0.876 & 0.729 & 0.919 & 0.796 & 0.800 \\
    w/- $L_{bind}$
        & 0.814 & 0.877 & 0.727 & 0.918 & 0.799 & 0.805\\
    \end{tabular}
}
%\vspace{-1.0em}
\end{center} 
%\vspace{-0.4em}
\caption{
    Ablation study on the binding loss $L_{bind}$. Despite the approach with the binding loss achieved similar performance or  minor improvement, we observed more accurate attribute localization as visualized in \cref{fig:app_visual_jsd}.
}
\label{tab:app_jsd_ablation}
%\vspace{-0.7em}
\end{table}

We ablate the effect of the proposed binding loss $L_{bind}$  qualitatively and quantitatively, as shown in \cref{fig:app_visual_jsd} and \cref{tab:app_jsd_ablation}. We observe that the binding loss introduce minor difference on the quantitative evaluation. We hypothesize that the coarse measurement of current evaluation metrics may not be able to reflect the advantage of our method and are not well aligned with human judgement\newcite{hu2023tifa,Lu2023LLMScoreUT}. As illustrated in \cref{fig:app_visual_jsd}, without the binding loss, the model is able to partially reflect the attribute but may mix with other attributes as well. For instance, in the second column, the front of the car is partially in green, which should be assigned to the balloon. While such imperfect results could still fool current evaluation metrics, as part of the car is indeed in pink. 
With $L_{bind}$, we can see the attributes can be more accurately localized at the corresponding object area. Therefore, we employ the binding loss by default, if the attributes are provided in the prompt.

%%%%%%%%%%%%%%%%%%%%%%%%%%%%%%%%%%%%%%%%%%%%
\section{Implementation \& Evaluation Details} \label{sec:appendix-eval}
%\vspace{-0.5em}
%% Algorithm %%
\renewcommand{\algorithmicrequire}{\textbf{Input:}}
\renewcommand{\algorithmicensure}{\textbf{Output:}}
\begin{center}
\begin{minipage}{0.75\textwidth}
    \begin{algorithm}[H]
    \caption{Simplified Algorithm Overview of {\ours}}\label{alg:algorithm}
    \begin{algorithmic}[1]
    \Require A text prompt $\mathcal{P}$ and a pretrained Stable Diffusion $SD$
    \Ensure A noised latent $z_{t-1}$ for the next denoising step
    \State Determine object $S$ and attribute $R$ tokens by GPT with in-context learning as in TIFA\newcite{hu2023tifa}
    %\For{$s \in S$}
    %\EndFor
    \State Extract attention maps for the object tokens $A_t^s$ and attribute tokens $A^r$
    \If{$A^r$ are not None}
        \State $L_{D\&B} = L_{attend} + \lambda L_{bind}$
    \Else
        \State $L_{D\&B} = L_{attend}$
    \EndIf
    \State $z_t' \gets z_t - \alpha_t \cdot  \nabla_{z_t} L_{D\&B}$
    \State $z_{t-1} \gets SD (z_t', \mathcal{P}, t)$
    \State  \Return $z_{t-1}$
    \end{algorithmic}
    \end{algorithm}
\end{minipage}
\end{center}

\paragraph{Algorithm Overview.}
We provide the algorithm overview in \cref{alg:algorithm}. Given the text prompt $\mathcal{P}$, we firstly identify the tokens of interest, e.g., object tokens and attribute tokens. This process can either be done manually or automatically with the aid of GPT-3\newcite{brown2020GPT}  as shown in \newcite{hu2023tifa}.
Taking advantage of the in-context learning\newcite{brown2020GPT,hu2022context} capability of GPT-3, by providing a few in-context examples, GPT-3 can automatically extract the desired nouns and adjectives for new input prompts. For instance, in our experiments on the COCO-Subject and COCO-Attribute benchmarks, we used the captions of COCO images without fixed templates as the prompts, where the object and attribute tokens were selected automatically using GPT-3.
Based on the token indices, we can extract attention maps and apply our $L_{B\&D}$ to update the noised latent $z_t$.

%%%%%%%%%%%%%%%%%%%%%%%%%%%%%%%%%%%%%%%
\paragraph{CLIP-Based Evaluation.}
For computing the CLIP-based similarity metrics, e.g., Text-Text similarity, Full Prompt Similarity and Minimum Object Similarity, we employ the pretrained CLIP VIT-B/16 model\newcite{radford2021clip}. To obtain the caption of generated images for Text-Text similarity evaluation, we use the BLIP\newcite{li2022blip} image captioning model finetuned on the MSCOCO Captions dataset\newcite{chen2015microsoft} from the LAVIS library
\newcite{li2022lavis}.

%%%%%%%%%%%%%%%%%%%%%%%%%%%%%%%%%%%%%%%
\paragraph{TIFA Evaluation.}
% TIFA Score
Evaluation of the TIFA metric is based on a performance of the visual-question-answering (VQA) system, e.g.~mPLUG\newcite{li-etal-2022-mplug}. By definition, the TIFA score is essentially the VQA accuracy.
Given the text input $\mathcal{T}$, we can generate $\mathcal{N}$ multiple-choice question-answer pairs $\{ Q_i, C_i, A_i\}_{i=1}^{N}$, where $Q_i$ is a question, $C_i$ is a set of possible choices and $A_i$ is the correct answer. These question-answer pairs can be designed manually or automatically produced by the large-scale language model, e.g.~GPT-3\newcite{brown2020GPT}. By providing a few in-context examples, GPT-3 can follow the instruction to generate question-answer pairs, and generalize to new text captions\newcite{hu2023tifa,hu2022incontext}.

%%%%%%%%%%%%%%%%%%%%%%%%%%%%%%%%%%%%%%%
\paragraph{Computational Complexity.}
Measured on a V100 GPU using 50 sampling steps, Stable
Diffusion takes approximately 13 seconds to generate a
single image. As we follow the hyperparameter settings as
Attend \& Excite\newcite{chefer2023attendandexcite}, both A\&E and our method have a similar average runtime of
17 seconds. The runtime slightly varies with the complexity
of prompts. 

\end{document}